\documentclass[11pt]{article}

\usepackage{acl}

\usepackage{times}
\usepackage{latexsym}
\usepackage[T1]{fontenc}
\usepackage[utf8]{inputenc}
\usepackage{microtype}
\usepackage{inconsolata}
\usepackage{graphicx}
\usepackage{booktabs}
\usepackage{amsfonts}
\usepackage{nicefrac}
\usepackage{xcolor}
\usepackage{caption}
\usepackage{amsmath}
\usepackage{wrapfig}
\usepackage{tabularx}
\usepackage{amsthm}
\usepackage{amssymb}
\usepackage{algorithm}
\usepackage{algorithmic}
\usepackage{pifont}
\usepackage{colortbl}
\usepackage[most]{tcolorbox}
\usepackage{enumitem}
\usepackage{longtable}

\newcommand{\sysname}{TTSR }
\newcommand{\github}{\raisebox{-1.5pt}{\includegraphics[height=1em]{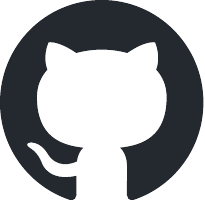}}}
\definecolor{rowblue}{HTML}{E6F3FF}
\definecolor{textred}{HTML}{FF3333}
\definecolor{textgreen}{HTML}{009900}

\title{TTSR: Test-Time Self-Evolving via Reflection}

\author{
  \textbf{Haoyang He\thanks{Equal contribution.}\textsuperscript{1}}, 
  \textbf{Zihua Rong\footnotemark[1]\textsuperscript{1}},\\
  \textbf{Yunjia Zhao\textsuperscript{2}},
  \textbf{Lan Yang\thanks{Corresponding author.}\textsuperscript{1}}, 
  \textbf{Jian Chang\textsuperscript{3}},
  \textbf{Honggang Zhang\textsuperscript{1}}
\\
  \textsuperscript{1}Beijing University of Posts and Telecommunications
\\
  \textsuperscript{2}Southwestern University of Finance and Economics
\\
  \textsuperscript{3}China Unicom Online Information Technology Co., Ltd.
\\
  \github\ Project: \href{https://github.com/Henryhe09/TTSR}{https://github.com/Henryhe09/TTSR}
}
\begin{document}
\maketitle

\begin{abstract}
Test-time training (TTT) adapts large language models (LLMs) during inference
using only unlabeled test inputs. Existing methods, however, face two major
bottlenecks on hard reasoning tasks: (1) \emph{lack of learnable samples}, as
self-generated pseudo-labels on difficult questions are often noisy and yield
unstable rewards; and (2) \emph{inefficient exploration}, as performance gains
depend on repeatedly sampling many rollouts without explicit diagnosis of why
previous attempts fail. We propose \textbf{TTSR}
(\textbf{T}est-\textbf{T}ime \textbf{S}elf-\textbf{R}eflection), a
self-evolving framework based on a \emph{reflect-then-synthesize} paradigm. A
single pretrained model alternates between a \textit{Student} role and a
\textit{Teacher} role: the Student solves test questions and updates,
while the Teacher analyzes failed trajectories and synthesizes targeted variant questions closer to the Student's
capability frontier. TTSR further
maintains a cross-iteration \textit{weakness memory} and compiles persistent
weaknesses into a lightweight \textit{strategy note} prepended to subsequent
Student inputs, so diagnostic knowledge can guide exploration and gradually
fade as weaknesses are resolved. Experiments on challenging mathematical
reasoning benchmarks show consistent test-time improvements, strong
cross-backbone generalization, and transfer to general-domain reasoning tasks.
\end{abstract}

\begin{figure*}[t]
    \centering
    
    \includegraphics[width=0.8\textwidth]{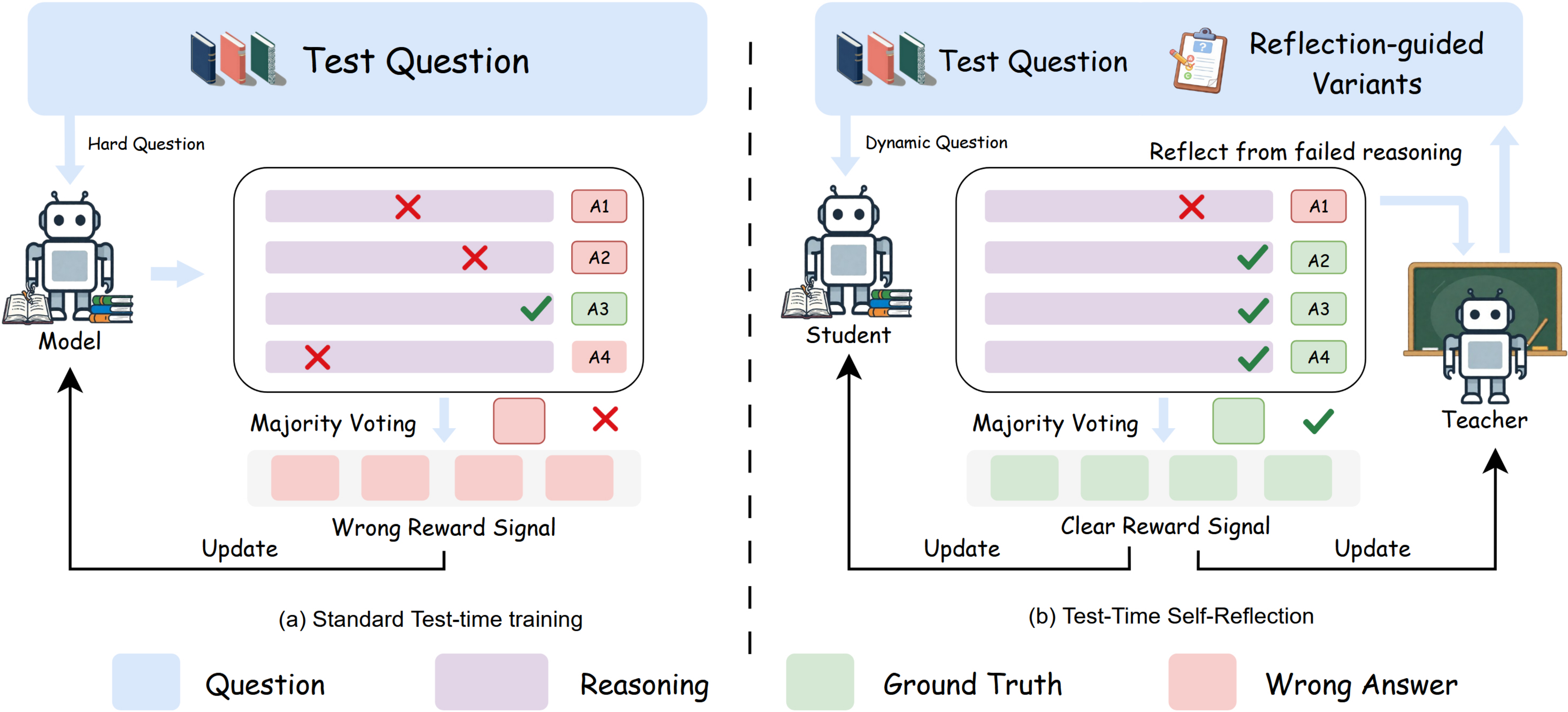}
    \caption{Comparison between standard Test-time training and TTSR. Standard test-time training produces noisy reward signals when test questions are too hard for the model. In contrast, TTSR reflects on failed reasoning to generate moderately difficult variants matched to the model's capability, yielding clearer rewards and more stable test-time updates.}
    \label{fig:overview}
\end{figure*}

\section{Introduction}

Test-time training (TTT) updates model parameters at inference time using only
unlabeled test data~\cite{akyurek2024surprising,sun2020testtimetrainingselfsupervisiongeneralization},
requiring no ground-truth supervision. Recent formulations cast TTT as
test-time reinforcement learning, where rewards are derived from the model's
own sampled solutions via self-consistency or outcome
verification~\cite{zuo2026ttrl,yuksekgonul2026learning}.
In practice, however, these methods produce limited and unstable gains on hard
reasoning tasks~\cite{huang2023large,akyurek2024surprising}. We focus on two
key limitations.

\textbf{Lack of learnable samples.}
Test questions in hard reasoning benchmarks often sit near or beyond what the
model can reliably solve. Majority-voting pseudo-labels then become noisy,
and the resulting rewards too unreliable for stable parameter
updates~\cite{akyurek2024surprising,zuo2026ttrl}.
In effect, only a narrow band of questions---neither trivially easy nor
hopelessly hard---yields useful gradients, leaving the model with few
effective training samples at test time~\cite{bae2025onlinedifficulty}.

\textbf{Inefficient exploration via repeated rollouts.}
In part because of this scarcity, current TTT methods sample large numbers of
reasoning trajectories, hoping that some will land on correct
solutions~\cite{snell2024scalingllmtesttimecompute}. This is expensive and
undirected: the model has no mechanism to diagnose \emph{why} earlier attempts
failed or to steer later sampling toward its actual weaknesses.

Two observations point toward a different approach. First, \emph{diagnosing
where a reasoning trace went wrong is typically easier than producing a correct
solution from scratch}, since error localization operates on a given trace
rather than requiring one to be
constructed~\cite{wang2022self,lin-etal-2024-criticbench,zheng2025processbench}. Second, curriculum
learning shows that training on calibrated intermediate problems improves
accuracy on harder target
tasks~\cite{bengio2009curriculum,wang2021surveycurriculumlearning}. Together,
these observations motivate a \emph{reflect-then-synthesize} strategy: the model
first diagnoses its reasoning failures, then uses the diagnoses to synthesize
variant questions near the model's capability frontier as stepping stones.

We propose \textbf{TTSR}
(\textbf{T}est-\textbf{T}ime \textbf{S}elf-\textbf{R}eflection), a test-time
self-evolving framework in which a \emph{single} pretrained model alternates
between two roles: a \emph{Student} and a \emph{Teacher}. The Student samples
reasoning trajectories for test questions and updates its policy online via
GRPO~\cite{shao2024deepseekmath} with self-supervised rewards. The Teacher
analyzes the Student's failed trajectories, extracts structured weakness
descriptors through lightweight reflection, and synthesizes variant questions
that target these weaknesses. The variants sit closer to the Student's
capability frontier and thus provide more informative training signals than the
original test questions. Per-iteration reflections are ephemeral: without persistent tracking, the same
weaknesses may surface repeatedly. TTSR therefore maintains a cross-iteration
\emph{weakness memory} and compiles the most persistent entries into a
\emph{strategy note}---concise reasoning guidelines prepended to Student inputs
that directly steer subsequent exploration toward unresolved weaknesses. The
note fades automatically as weaknesses are resolved through continued
adaptation.

We evaluate TTSR on eight mathematical and general reasoning benchmarks across
three model backbones. TTSR consistently outperforms test-time RL baselines,
achieving up to $+15.6$ average points over the base model, and transfers to
general-domain reasoning tasks without domain-specific tuning.
These results suggest that \emph{self-reflection with cross-iteration knowledge
accumulation} provides a practical pathway for continual reasoning
improvement during inference.

\paragraph{Contributions.}
Our main contributions are:
\begin{itemize}
    \item We identify the lack of learnable samples and inefficient
    rollout-based exploration as two key bottlenecks of test-time training, and
    propose a \emph{reflect-then-synthesize} strategy that leverages failure
    diagnosis to construct targeted curricula near the model's capability
    frontier.
    \item We introduce \textbf{TTSR}, a test-time self-evolving framework in
    which a single model alternates between Student and Teacher roles. TTSR
    further incorporates a cross-iteration \emph{weakness memory} and a
    lightweight \emph{strategy note} that persistently carry forward diagnostic
    knowledge across adaptation iterations.
    \item We demonstrate that TTSR consistently improves performance on challenging mathematical and general reasoning benchmarks and generalizes across model
    backbones.
\end{itemize}

\section{Related Work}

\paragraph{Self-Play and Curriculum Learning for LLMs (mainly training-time).}
Self-play has become a common recipe for autonomous improvement in LLMs, from iterative self-training and alignment-oriented loops to role-specialized or co-evolutionary variants~\cite{huang2022largelanguagemodelsselfimprove,selfinstruct,chen2024self,lin2025learningsolveverifyselfplay,chen2025spcevolvingselfplaycritic,fang2025serlselfplayreinforcementlearning}. Recent capability-oriented systems further emphasize adversarial generation, dynamic self-challenging, and data-free evolution~\cite{cheng2025spice,xu2025genius,wang2025octothinkermidtrainingincentivizesreinforcement,kuba2025languageselfplay,zhou2026self,wei2025sftsecondrlupt,huang2025rzeroselfevolvingreasoningllm,zhao2026absolute,he2025visplay}. In parallel, curriculum learning establishes that calibrated intermediate tasks can improve learning efficiency~\cite{bengio2009curriculum,wang2021surveycurriculumlearning}. However, most of this line is designed for training/post-training regimes with larger budgets and does not explicitly target per-instance failure diagnosis during inference.

\paragraph{Test-Time Training and Test-Time RL.}
Test-time training (TTT) adapts model parameters at inference time using unlabeled test inputs~\cite{sun2020testtimetrainingselfsupervisiongeneralization,akyurek2024surprising}. TTRL-style methods cast this as test-time reinforcement learning with self-generated rewards from voting or verification~\cite{zuo2026ttrl,yuksekgonul2026learning}. TTCS introduces curriculum learning into test-time training, but its synthesized questions are derived from the original test questions rather than explicit diagnoses of the model's reasoning failures~\cite{yang2026ttcs}. These methods show that online adaptation is feasible, but two challenges remain on hard reasoning benchmarks: pseudo-label reliability degrades when questions are far beyond current capability, and large rollout budgets are often used as a coarse exploration mechanism~\cite{snell2024scalingllmtesttimecompute,bae2025onlinedifficulty}.

\paragraph{Positioning of TTSR.}
TTSR is a failure-conditioned curriculum adaptation mechanism for test-time training. Its first contribution is that we introduce self-generated curriculum construction into the test-time stage, where adaptation must be online and label-free, while difficult test inputs often lead to too few learnable samples and unstable updates in TTRL. Its second contribution is a \emph{reflect-then-synthesize} loop that converts failed reasoning traces into targeted variant questions near the model's current capability frontier, improving synthesis and exploration efficiency.

\section{TTSR: Test-Time Self-Reflection}
\label{sec:method}

\begin{figure*}[t]
    \centering
    \includegraphics[width=0.83\textwidth]{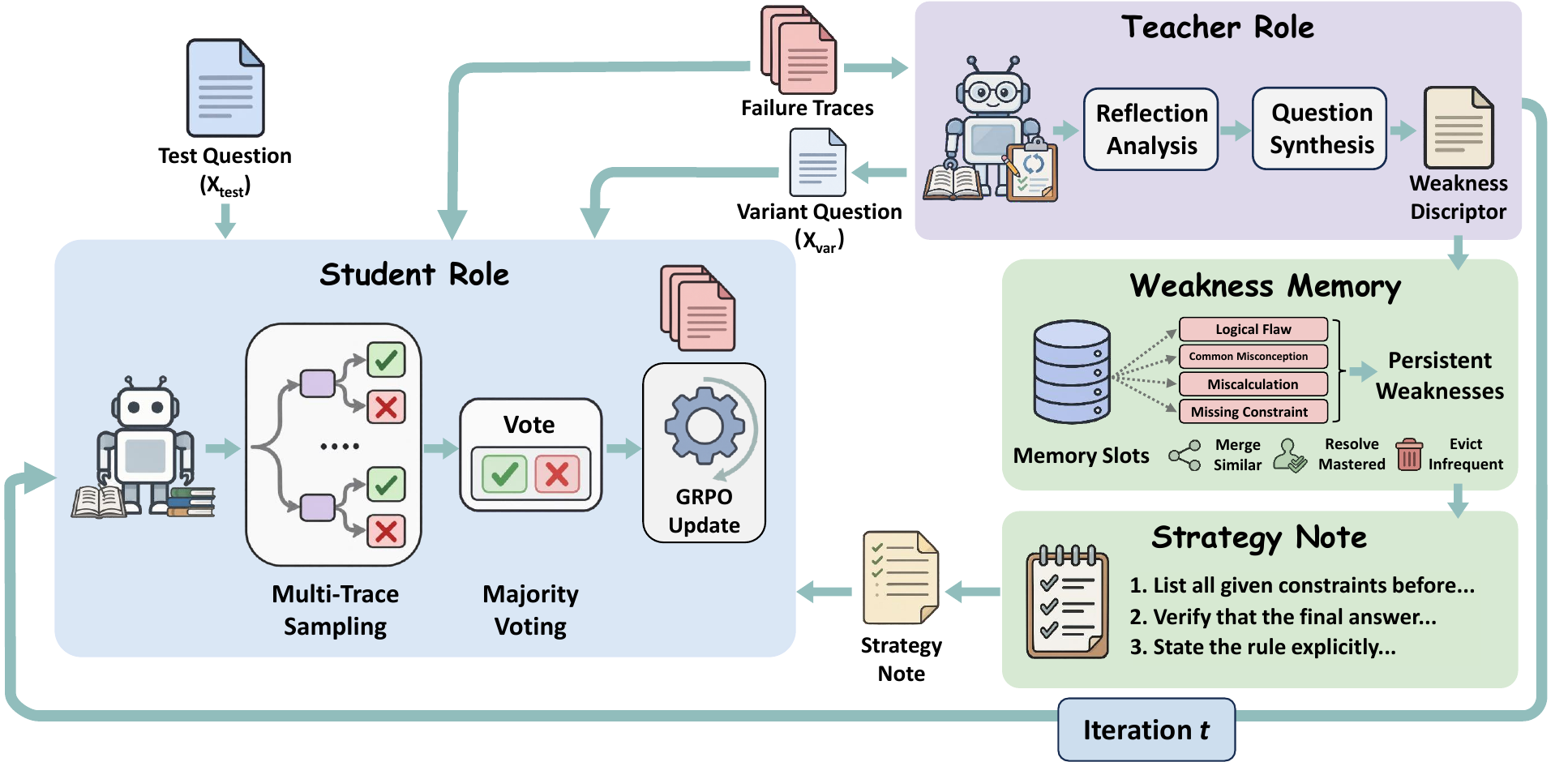}
    \caption{\textbf{Overview of the TTSR pipeline.} At each iteration $t$, the Student samples multiple reasoning trajectories and updates its policy via GRPO with majority-voting pseudo-labels. The Teacher analyzes failed traces to extract structured weakness descriptors, which are accumulated in a cross-iteration \emph{weakness memory}. Persistent weaknesses are compiled into a \emph{strategy note} that is prepended to Student inputs, providing immediate reasoning guidance. The Teacher also synthesizes targeted variant questions conditioned on both current failures and the accumulated memory, forming a closed-loop self-evolving cycle.}
    \label{fig:pipeline}
\end{figure*}

We present \textbf{TTSR}, a test-time self-evolving framework built on GRPO
and TTT. A single pretrained model $\pi_\theta$ alternates between two roles
at test time: a \emph{Student} that solves test questions and updates its
policy via self-supervised rewards, and a \emph{Teacher} that analyzes failed
reasoning trajectories, extracts structured weakness descriptors, and
synthesizes targeted variant questions. To accumulate diagnostic knowledge
across iterations, TTSR maintains a \emph{weakness memory} $\mathcal{M}_t$
and compiles persistent entries into a \emph{strategy note}
$h_t$ that provides concise reasoning guidelines prepended to Student inputs.
The overall pipeline is illustrated in Figure~\ref{fig:pipeline}.

\paragraph{Student Role.}
Given a test question $x \in \mathcal{X}_{\text{test}}$ and the current
strategy note $h_t$ (Section~\ref{subsec:memory}), the Student samples $G$
reasoning trajectories
$\{y_i\}_{i=1}^G \sim \pi_\theta(\cdot \mid h_t,\, x)$,
each consisting of a chain-of-thought process and a final answer.
We write $a(y_i)$ for the final answer extracted from trajectory $y_i$.
These trajectories serve both as inference outputs and as the basis for
self-supervised learning signals.

\paragraph{Teacher Role.}
The Teacher does not solve the test question directly.
Instead, it identifies failures in the Student's trajectories relative to a
pseudo-target obtained via majority voting, reflects on the underlying
reasoning weaknesses, and synthesizes targeted easier variant questions
designed to expose and correct them.

\subsection{Student: Solving Problems and Test-Time Adaptation}
\label{subsec:student}

\paragraph{Test-Time Training Set Construction.}
At test-time iteration $t$, the Student updates its policy $\pi_{\theta_t}$ using a dynamically constructed training set.
Specifically, we define the training dataset at iteration $t$ as
\begin{equation}
\mathcal{D}_t
=
\mathcal{X}_{\text{test}}
\;\cup\;
\mathcal{X}^{(t-1)}_{\text{var}},
\end{equation}
where $\mathcal{X}_{\text{test}}$ denotes the original test set, and $\mathcal{X}^{(t-1)}_{\text{var}}$ represents the set of variant questions synthesized by the Teacher using the previous policy $\pi_{\theta_{t-1}}$.
The Student samples training questions from $\mathcal{D}_t$ and performs test-time adaptation via reward-based optimization.

\paragraph{Majority Voting Reward.}
For each training question $x \in \mathcal{D}_t$, the Student samples a group of $G$ reasoning trajectories using the current policy:
\begin{equation}
\{y_i\}_{i=1}^{G} \sim \pi_{\theta_t}(\cdot \mid h_t,\, x).
\end{equation}
Following the Test-Time Reinforcement Learning paradigm, we aggregate the sampled trajectories via majority voting over their final answers to obtain a consensus answer $\hat{a}(x)$, which serves as a pseudo-correct reference.
An outcome-level pseudo-correctness reward is then assigned to each trajectory according to whether its final answer agrees with the consensus:
\begin{equation}\label{eq:pseudo-correctness reward}
R_S(y_i \mid x) =
\mathbb{I}\!\left[a(y_i) = \hat{a}(x)\right].
\end{equation}

Based on these rewards, we define the empirical pseudo-correctness score for question $x$ as
\begin{equation}\label{eq:empirical pseudo-correctness score}
s_t(x) = \frac{1}{G}\sum_{i=1}^{G} R_S(y_i \mid x),
\end{equation}
which measures the stability of the Student's predictions under stochastic sampling at iteration $t$.
The Student is optimized using GRPO with these pseudo-correctness rewards, enabling stable and label-free test-time adaptation.

\subsection{Teacher: Reflection-Guided Curriculum Synthesis}
\label{subsec:teacher}
\paragraph{Reflection on Reasoning Steps.}
At test-time iteration $t$, the Teacher performs reflection based on the Student's failed reasoning behaviors observed in the previous iteration.
Specifically, using the policy $\pi_{\theta_{t-1}}$, we collect a set of reasoning trajectories whose pseudo-correctness rewards are zero.
From these trajectories, we randomly sample a subset of failed instances for reflection.

Each sampled instance consists of a tuple $(x, y, \hat{a})$, where $x$ denotes the question, $y$ the Student's reasoning trajectory, and $\hat{a}$ the corresponding pseudo-correct answer obtained via majority voting over final answers.
Conditioned on this set of failed tuples, the Teacher with policy $\pi_{\theta_t}$ analyzes the reasoning steps in $y$ and produces a reflection summary that characterizes both the incorrect reasoning patterns and the missing or insufficient steps leading to failure.
This reflection focuses on the process-level deficiencies of the Student's reasoning rather than the final answer correctness.
Each reflection output is a structured weakness descriptor consisting of a natural-language weakness summary, trigger conditions, and failure signatures.
These descriptors serve dual purposes: guiding question synthesis within the current iteration and updating the cross-iteration weakness memory $\mathcal{M}_t$ (Section~\ref{subsec:memory}).

\paragraph{Reflection-Guided Question Synthesis.}
Given the extracted weakness descriptors and the accumulated weakness memory,
the Teacher synthesizes variant questions that form an adaptive curriculum for
the Student. Let $\mathcal{F}^{(t-1)} = \{(x_k, y_k, \hat{a}_k)\}_{k=1}^{|\mathcal{F}|}$
denote the set of failed instances from the previous iteration. We construct a
synthesis prompt $p_t$ from both $\mathcal{F}^{(t-1)}$ and the most persistent
entries in $\mathcal{M}_t$, then sample $M$ variants:
\begin{equation}
\mathcal{X}^{(t)}_{\text{var}}
=
\left\{ x'_j \right\}_{j=1}^{M}
\sim
\pi_{\theta_t}\!\left(
\cdot
\;\middle|\;
p_t,\,
\mathcal{F}^{(t-1)},\,
\mathcal{M}_t
\right).
\end{equation}
Each variant preserves the core reasoning structure of the original question
while modifying conditions to expose the identified weaknesses. We use group
sampling to encourage diversity, so the curriculum covers multiple
manifestations of the same deficiency.

\paragraph{Difficulty Reward.}
We evaluate the difficulty of a variant question $x'$ using the Student's pseudo-correctness score at iteration $t$.
Concretely, the Student samples $G$ trajectories under $\pi_{\theta_t}$ and computes the empirical pseudo-correctness score $s_t(x') \in [0,1]$ as defined in Section~\ref{subsec:student}.

Following prior work showing that variance-based rewards effectively induce capability-centered curricula in self-play~\cite{cheng2025spice}, we adopt a normalized variance-based frontier reward:
\begin{equation}
R_{\text{diff}}(x')
=
4\, s_t(x')\bigl(1 - s_t(x')\bigr).
\end{equation}

This reward lies in $[0,1]$ and is maximized at $s_t(x') = 0.5$, i.e., when the Student operates near its capability frontier.
It therefore encourages the Teacher to synthesize variant questions that are neither trivial nor degenerate.
Further theoretical analysis is provided in Appendix~\ref{app:difficulty_reward_theory}.
\paragraph{Similarity Penalty.}
\label{method:spr}
To reduce redundancy among synthesized variants, we penalize near-duplicate
generations. Let $\mathcal{Z} = \{x'_1,\dots,x'_M\} \cup \{x\}$ be the set
of variants together with the original question. For each variant $x'_i$, the
penalty is
\begin{equation}
\begin{aligned}
&P_{\text{sim}}(x'_i, x)
=
\frac{1}{|\mathcal{Z}|-1}\\
&\sum_{z \in \mathcal{Z}\setminus\{x'_i\}}
\max\!\left(0,\ \mathrm{sim}(x'_i,z)-\tau\right).
\end{aligned}
\end{equation}
where $\mathrm{sim}(\cdot,\cdot) \in [0,1]$ is a token-level sequence
similarity metric (Appendix~\ref{app:similarity_metric}) and $\tau$ controls
the overlap tolerance.

\paragraph{Teacher Reward.}
For each variant $x'_i$, the Teacher's overall reward combines difficulty and
diversity:
\begin{equation}
\begin{aligned}
R_T(x'_i)
&=
\max\!\left(
0,\;
R_{\text{diff}}(x'_i)
-
\lambda\,P_{\text{sim}}(x'_i, x)
\right).
\end{aligned}
\end{equation}
Outputs that violate the predefined \texttt{<question>} format are discarded
and receive zero reward.

\subsection{Cross-Iteration Weakness Memory and Strategy Note}
\label{subsec:memory}

Per-iteration reflections are ephemeral and do not accumulate knowledge across
iterations. To persistently track reasoning deficiencies, TTSR maintains a
\emph{weakness memory} $\mathcal{M}_t$ and compiles persistent entries into a
\emph{strategy note} $h_t$.

\paragraph{Weakness Memory.}
At each iteration $t$, the Teacher's reflection produces weakness descriptors
$\mathcal{W}^{(t)} = \{w^{(t)}_1, \dots, w^{(t)}_n\}$
(Section~\ref{subsec:teacher}). The memory $\mathcal{M}_t$ is updated via
three operations:

\textbf{Merge.}
If a new descriptor $w' \in \mathcal{W}^{(t)}$ matches an existing entry
$w \in \mathcal{M}_{t-1}$ ($\mathrm{sim}(w', w) > \tau_w$, using the same
metric as in Section~\ref{method:spr}), we merge $w'$ into $w$ by
incrementing its persistence counter $c(w) \leftarrow c(w) + 1$ and updating
$w.\mathrm{last\_seen} \leftarrow t$; otherwise $w'$ is added as a new entry
with $c(w') = 1$.

\textbf{Resolve.}
Entries not observed for $\Delta$ consecutive iterations
($t - w.\mathrm{last\_seen} > \Delta$) are removed, indicating the weakness
has been addressed.

\textbf{Evict.}
If $|\mathcal{M}_t| > K$ after merging, entries with the lowest persistence
count are dropped.

The complete update is:
\begin{equation}
\begin{aligned}
\widetilde{\mathcal{M}}_{t+1}
&=
\mathrm{Resolve}\!\big(
\mathrm{Merge}(\mathcal{M}_t,\, \mathcal{W}^{(t)}),\, \Delta
\big),
\\
\mathcal{M}_{t+1}
&=
\mathrm{Evict}\!\big(\widetilde{\mathcal{M}}_{t+1},\, K\big).
\end{aligned}
\end{equation}

\paragraph{Strategy Note.}
At the beginning of each iteration, we select the top-$N$ entries from
$\mathcal{M}_t$ ranked by persistence count and compile them into a strategy
note:
\begin{equation}
h_t = \mathrm{Compile}(\mathcal{M}_t^{\,\mathrm{top}\text{-}N}).
\end{equation}
The note is prepended to all Student inputs during both sampling and GRPO
training. As the Student improves, resolved weaknesses are removed from
$\mathcal{M}_t$ and the corresponding guidelines automatically drop from
$h_t$, so the note serves as a transient scaffold that fades as the
underlying capability catches up.

\section{Experiment}
\subsection{Experimental Setting}
\paragraph{Models and Baselines.}
We evaluate on three base pretrained language models:
\textbf{Qwen3-4B-Base}, \textbf{Qwen3-8B-Base}~\cite{yang2025qwen3}, and \textbf{OctoThinker-8B-Hybrid-Base}~\cite{wang2025octothinkermidtrainingincentivizesreinforcement}.
We compare against two test-time optimization baselines:
\textbf{TTRL}~\cite{zuo2026ttrl}, which adapts the model on unlabeled test instances using self-generated reinforcement signals, and \textbf{R-Zero}~\cite{huang2025rzeroselfevolvingreasoningllm}, which enables data-free self-improvement through reinforcement learning.
The \textbf{Base model} performs direct inference without parameter updates.
We additionally report a \textbf{Self-Consistency} (SC)~\cite{wang2022self} baseline that applies majority voting over $G\!=\!16$ sampled trajectories without any parameter updates, to disentangle the effect of inference-time scaling from test-time training.
All baselines are evaluated under the same test-time setting for fair comparison.

\paragraph{Datasets and Evaluation.}
We evaluate mathematical reasoning on \textbf{AMC23}, \textbf{MATH-500}~\cite{MATH-500}, \textbf{Minerva}~\cite{Minerva}, \textbf{OlympiadBench}~\cite{OlympiadBench}, \textbf{AIME 2024}, and \textbf{AIME 2025}~\cite{li2025limr}, which cover a wide range of difficulty levels and require multi-step symbolic manipulation.
We further evaluate general reasoning on \textbf{GPQA-Diamond}~\cite{gpqa} and \textbf{MMLU-Pro}~\cite{wang2024mmluprorobustchallengingmultitask} to assess reasoning ability beyond mathematics.
Each benchmark is adapted independently.

\paragraph{Training Configuration.}
The Teacher generates $M\!=\!2$ variant questions per original test question per iteration; the Student samples $G\!=\!16$ reasoning trajectories per question.
The policy is optimized via GRPO with pseudo-correctness rewards.
All hyperparameters, including similarity penalty ($\lambda\!=\!1.0$, $\tau\!=\!0.75$), weakness memory ($K\!=\!10$, $\Delta\!=\!3$, $\tau_w\!=\!0.6$) and strategy note ($N\!=\!3$), are fixed across all models and benchmarks; see Appendix~\ref{app:training_setting} for full details.

\begin{table*}[t]
\centering
\small
\renewcommand{\arraystretch}{1.05}
\setlength{\tabcolsep}{1.5pt}
\begin{tabular}{@{}lccccccccc@{}}
\toprule
\textbf{Method} & \textbf{AMC23} & \textbf{MATH500} & \textbf{Minerva} & \textbf{Olympiad} & \textbf{AIME24} & \textbf{AIME25} & \textbf{GPQA-D} & \textbf{MMLU-Pro} & \textbf{$\Delta$} \\
\midrule
\multicolumn{10}{@{}l}{\cellcolor{cyan!6}\textit{Qwen3-4B-Base}} \\
Base Model & 45.0 & 72.0 & 32.4 & 40.2 & 12.4 & 5.8 & 25.8 & 52.0 & -- \\
R-Zero & 55.0 & 76.8 & 40.8 & 44.1 & 18.2 & 9.1 & 28.3 & 55.0 & +5.2 \\
TTRL & 55.0 & 79.0 & 43.8 & 46.0 & 17.6 & 14.7 & 29.3 & 56.0 & +7.0 \\
\rowcolor{yellow!12}
\textbf{TTSR (ours)} & \textbf{60.0} & \textbf{82.4} & \textbf{52.9} & \textbf{45.3} & \textbf{25.6} & \textbf{17.6} & \textbf{34.3} & \textbf{60.8} & \textbf{+11.7} \\
\midrule
\multicolumn{10}{@{}l}{\cellcolor{cyan!6}\textit{Qwen3-8B-Base}} \\
Base Model & 52.5 & 77.8 & 39.7 & 41.2 & 15.9 & 9.8 & 33.3 & 58.6 & -- \\
R-Zero & 57.5 & 82.0 & 47.8 & 48.7 & 22.6 & 13.4 & 36.9 & 61.5 & +5.2 \\
TTRL & 62.5 & 84.2 & 50.4 & 50.7 & 26.1 & 15.7 & 38.4 & 62.8 & +7.8 \\
\rowcolor{yellow!12}
\textbf{TTSR (ours)} & \textbf{67.5} & \textbf{87.4} & \textbf{54.8} & \textbf{55.2} & \textbf{30.8} & \textbf{19.1} & \textbf{42.4} & \textbf{66.7} & \textbf{+11.9} \\
\midrule
\multicolumn{10}{@{}l}{\cellcolor{cyan!6}\textit{OctoThinker-8B-Hybrid-Base}} \\
Base Model & 27.5 & 45.6 & 15.1 & 16.8 & 7.1 & 3.9 & 15.2 & 26.8 & -- \\
R-Zero & 35.0 & 54.0 & 22.8 & 25.4 & 11.6 & 6.8 & 19.7 & 31.4 & +6.1 \\
TTRL & 37.5 & 57.2 & 26.5 & 28.6 & 13.9 & 8.2 & 21.7 & 33.2 & +8.6 \\
\rowcolor{yellow!12}
\textbf{TTSR (ours)} & \textbf{47.5} & \textbf{64.8} & \textbf{34.2} & \textbf{36.8} & \textbf{19.7} & \textbf{12.4} & \textbf{27.8} & \textbf{39.8} & \textbf{+15.6} \\
\bottomrule
\end{tabular}
\caption{Comprehensive evaluation across mathematical and general reasoning benchmarks. AIME 2024/2025 report Mean@32 accuracy; all other benchmarks report Pass@1 (greedy decoding).}
\label{tab:main_results}
\end{table*}

\subsection{Main Results}

Table~\ref{tab:main_results} summarizes results across all benchmarks.

\textbf{\sysname achieves consistent improvements across models and domains.}
On Qwen3-4B-Base, \sysname raises average accuracy by $+11.7$ points over the base model.
On Qwen3-8B-Base, the average improvement is $+11.9$ points, and on OctoThinker-8B-Hybrid-Base, which undergoes reasoning-oriented mid-training~\cite{wang2025octothinkermidtrainingincentivizesreinforcement}, the gains reach $+15.6$ points.
Beyond mathematics, \sysname also improves general reasoning: gains on GPQA-Diamond and MMLU-Pro range from $3$ to $7$ points over the strongest baseline, indicating that the reflection-guided curriculum transfers across domains.

\textbf{\sysname outperforms existing test-time training methods on average and on most benchmarks.}
Compared to TTRL and R-Zero, \sysname delivers clear average gains across model backbones, with improvements on the majority of benchmarks.
On Qwen3-8B-Base, \sysname surpasses TTRL by $+4.1$ points on average; on OctoThinker-8B, the margin widens to $+7.0$.
This gap indicates that reflection-guided variant synthesis provides more informative training signal than direct adaptation on raw test questions.

\textbf{\sysname shows the largest improvements on challenging benchmarks.}
The advantage is most pronounced on AIME 2024 and AIME 2025, where the original questions lie far beyond the model's reliable solving range.
On Qwen3-4B-Base, \sysname exceeds TTRL by $+8.0$ on AIME24 and $+2.9$ on AIME25.
On Qwen3-8B-Base, \sysname reaches $30.8$ on AIME24 ($+4.7$ over TTRL) and $19.1$ on AIME25 ($+3.4$).
When TTRL's pseudo-labels become unreliable on intractable problems, Teacher-synthesized variants bridge the difficulty gap by creating frontier-level stepping stones that progressively approach the target difficulty.

We additionally compare against Self-Consistency (SC), which applies majority voting over $G\!=\!16$ trajectories without parameter updates; full results are reported in Table~\ref{tab:sc_comparison} (Appendix~\ref{app:sc_comparison}). All test-time training methods substantially outperform SC, confirming that \sysname's gains stem from parameter adaptation rather than inference-time scaling.

\subsection{Analysis}

We conduct targeted analyses to validate the key design choices of TTSR.

\paragraph{Q1: Do synthesized variants concentrate near the capability frontier?}
A key hypothesis is that reflection-guided synthesis generates questions near the Student's capability frontier, where $R_{\text{diff}} = 4s_t(1-s_t)$ is high.
On Qwen3-8B-Base, we compare pseudo-correctness $s_t(x')$ across three sources: (i) original test questions, (ii) variants without reflection (random structural variation), and (iii) full TTSR variants.
We report the \emph{frontier ratio}: the fraction with $s_t \in [0.2, 0.8]$, where $R_{\text{diff}}$ is at least $64\%$ of its maximum (Table~\ref{tab:variant_quality}, Appendix~\ref{app:variant_quality}).
On MATH500, the frontier ratio rises from $16.2\%$ (original) to $67.4\%$ (TTSR), a $4.2\times$ increase.
On AIME25, where original questions have a frontier ratio of only $8.3\%$, TTSR variants achieve $64.2\%$, demonstrating that reflection is especially valuable when the test set lies far from the model's capability boundary.
Removing reflection reduces the frontier ratio to $43.8\%$ and $35.6\%$, indicating that weakness-aware diagnosis provides gains beyond what random structural variation can achieve.

\paragraph{Q2: Is co-evolution necessary, or can a stronger static teacher suffice?}
The ablation study (Table~\ref{tab:ablation_full}, Appendix~\ref{app:ablation_full}) shows that removing Teacher test-time updates (equivalent to using a frozen
Teacher at iteration 0) causes the largest overall drop (up to $-5.8$ on AIME25 and Olympiad).
This confirms that Teacher--Student co-evolution is critical: a co-evolving same-size Teacher keeps the curriculum aligned with the Student's evolving capability frontier, whereas a frozen Teacher produces increasingly mismatched variants as training progresses.

\paragraph{Q3: How does TTSR perform with limited test data?}
To assess whether TTSR's gains rely on large test sets, we evaluate data
efficiency on \textbf{AIME24} with \textbf{Qwen3-4B-Base}. TTSR consistently
outperforms TTRL at all data ratios: $10\%$ ($12.3$ vs $9.5$, $+2.8$), $20\%$
($14.6$ vs $11.2$, $+3.4$), $30\%$ ($15.3$ vs $11.8$, $+3.5$), and $100\%$
($25.6$ vs $17.6$, $+8.0$). These results suggest that reflection-guided
variant synthesis amplifies the value of limited test data by constructing
additional frontier-difficulty training signals. It also weakens an
instance-memorization explanation: even with only a small fraction of test
instances, TTSR retains a clear advantage.

\begin{figure}[t]
\centering
\includegraphics[width=\linewidth]{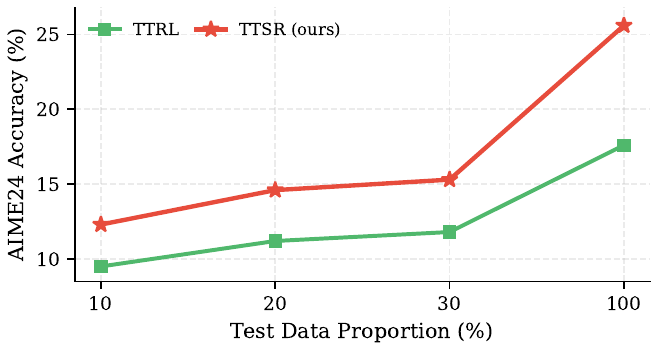}
\caption{\textbf{Data efficiency on AIME24 (Qwen3-4B-Base).}
Accuracy under four test-data proportions.}
\label{fig:data_efficiency}
\end{figure}

Appendix~\ref{app:compute} further analyzes the computational cost of TTSR through a per-iteration breakdown and an iso-compute comparison against an extended TTRL run. The matched-budget experiment shows that TTRL largely saturates with additional iterations, whereas TTSR retains a clear advantage. This result supports that the gains arise from the reflection-guided curriculum rather than additional computation alone.

\paragraph{Q4: Do improvements transfer across domains?}
We evaluate cross-benchmark transfer on Qwen3-8B-Base.
The OOD experiments (Figure~\ref{fig:ood_performance}) show that TTSR transfers well across mathematical benchmarks: training on Olympiad yields $25.4$ on AIME24, exceeding TTRL's $22.1$ ($+3.3$), while the Base model achieves only $15.9$.
Beyond mathematics, the training dynamics (Figure~\ref{fig:training_dynamics}) reveal meaningful cross-domain gains on GPQA-D ($33.1 \!\rightarrow\! 36.2$, $+3.1$) and MMLU-Pro ($58.6 \!\rightarrow\! 61.2$, $+2.6$), far exceeding TTRL ($+1.5$ and $+1.2$, respectively).
These cross-domain gains suggest that TTSR acquires transferable reasoning strategies rather than memorizing benchmark-specific patterns.

\subsection{Training Dynamics}

\begin{figure*}[t]
    \centering
    \includegraphics[width=\linewidth]{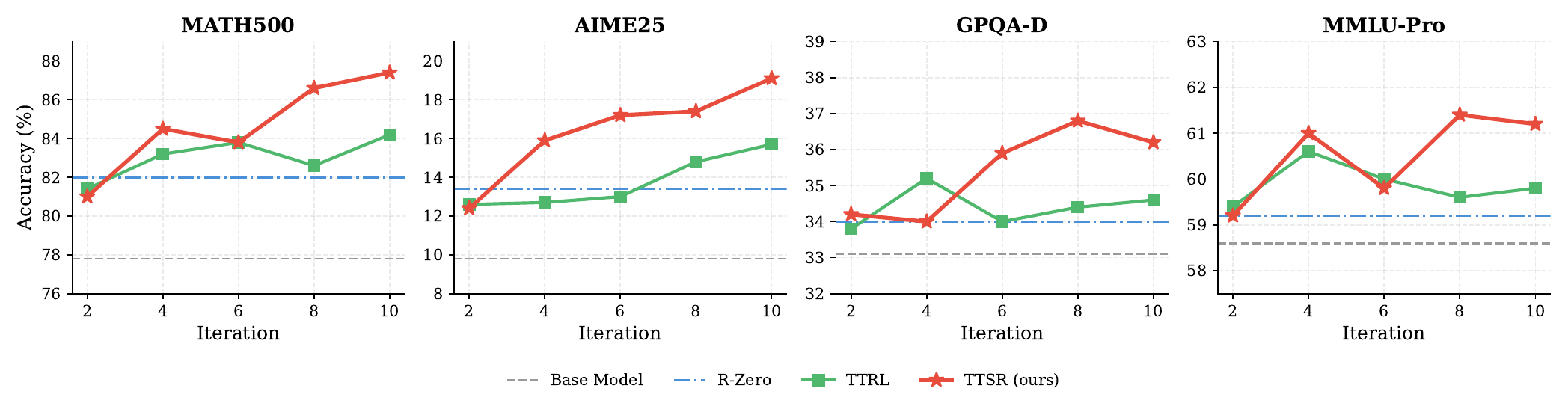}
    \caption{
    \textbf{Training dynamics of TTSR on Qwen3-8B-Base.}
    Left two panels show \emph{in-domain} accuracy on the training benchmarks (MATH500, AIME25); right two panels show \emph{cross-domain} accuracy on held-out general reasoning benchmarks (GPQA-D, MMLU-Pro) while training on AIME25. Dashed and dash-dotted horizontal lines denote the Base Model and R-Zero baselines, respectively.
    }
    \label{fig:training_dynamics}
\end{figure*}

Figure~\ref{fig:training_dynamics} tracks accuracy across test-time iterations on Qwen3-8B-Base.
On both MATH500 and AIME25, TTSR starts below TTRL at iteration~2 but overtakes TTRL in later iterations.
TTRL shows larger intermediate fluctuations, including a dip on MATH500 at iteration~8 ($83.8 \!\to\! 82.6$), whereas TTSR exhibits steadier upward progress.
Even when trained exclusively on AIME25, TTSR yields cross-domain gains on GPQA-D ($+3.1$) and MMLU-Pro ($+2.6$), exceeding TTRL ($+1.5$ and $+1.2$), demonstrating that the gains transfer beyond the mathematical domain, and all baselines share identical access to test data, so comparisons remain fair.
Additional training dynamics on AMC23 and Olympiad are provided in Appendix~\ref{app:training_dynamics}.

\subsection{Ablation Study}

Table~\ref{tab:ablation} progressively enriches the variant-synthesis pipeline on Qwen3-8B-Base to isolate the source of TTSR's gains.
A full component-level ablation is deferred to Table~\ref{tab:ablation_full} (Appendix~\ref{app:ablation_full}).

\textbf{Variant synthesis already helps.}
Adding difficulty-controlled variant generation with a same-size (8B) teacher and training via GRPO (+\,Synthesis) lifts accuracy by $+2.6$--$+6.4$ points over the base model, confirming that synthesized curriculum data provides meaningful learning signals beyond the original test set.

\textbf{Scaling the teacher has diminishing returns.}
Replacing the 8B teacher with Qwen3-14B (+\,Strong Synthesizer) yields an additional $+1.5$--$+3.4$ points.
While the larger teacher generates higher-quality variants, the gains plateau because, without reflection and co-evolution, the curriculum cannot adapt to the Student's evolving weaknesses.

\textbf{TTSR's framework matters most.}
TTSR adds reflection-guided synthesis, Teacher--Student co-evolution, and auxiliary components on top of the same-size 8B teacher, achieving a further $+2.8$--$+5.0$ points over the 14B synthesizer.
The total improvement of TTSR over the base model ($+9.1$--$+14.0$) is $1.6\times$ larger than the strong synthesizer's ($+4.1$--$+9.8$), demonstrating that TTSR's adaptive curriculum design matters more than raw teacher capacity.

\begin{table}[t]
\centering
\setlength{\tabcolsep}{4pt}
\renewcommand{\arraystretch}{1.15}

\begin{tabular}{@{}lcccc@{}}
\toprule
\textbf{Config.} & \textbf{M500} & \textbf{A25} & \textbf{Oly.} & \textbf{GPQA-D} \\
\midrule
Base Model
& 77.8
& 9.8
& 41.2
& 33.3 \\

+ Synthesis
& 82.2
& 12.8
& 47.6
& 35.9 \\

+ Strong Synth.
& 84.6
& 15.0
& 51.0
& 37.4 \\

\rowcolor{yellow!12}
+ \textbf{TTSR (Ours)}
& \textbf{87.4}
& \textbf{19.1}
& \textbf{55.2}
& \textbf{42.4} \\

\bottomrule
\end{tabular}
\caption{Additive analysis on Qwen3-8B-Base. Starting from the base model, we progressively enrich the variant-synthesis pipeline. ``Strong Synthesizer'' replaces the same-size teacher. Full component-level ablation is in Table~\ref{tab:ablation_full} (Appendix~\ref{app:ablation_full}).}
\label{tab:ablation}
\end{table}

\section{Conclusion}

We identify two core bottlenecks of test-time training on hard reasoning tasks: noisy pseudo-labels on overly difficult questions (few learnable samples) and inefficient rollout exploration without failure diagnosis. To address both, we propose \textbf{TTSR}, a \emph{reflect-then-synthesize} framework in which one model alternates between Student and Teacher roles. The Teacher analyzes failed trajectories, extracts structured weaknesses, and synthesizes targeted variants near the Student's capability frontier. A cross-iteration weakness memory compiles persistent patterns into a strategy note that guides subsequent reasoning.

Across eight mathematical and general reasoning benchmarks and three base models, TTSR improves average accuracy by $+11.7$ to $+15.6$ points over direct inference and consistently outperforms prior test-time training baselines. Ablations show that Teacher--Student co-evolution and reflection-guided synthesis are the primary contributors, and cross-domain results indicate that gains transfer beyond the training distribution. Training dynamics and cross-domain evaluations further demonstrate sustained improvement across iterations and transfer to held-out reasoning domains.

\section*{Limitations}

In this work, we introduce TTSR, a test-time self-evolving framework that improves large language models on challenging reasoning tasks through reflection-guided curriculum synthesis and cross-iteration weakness accumulation. However, there are still some limitations:
(1) Our current evaluation mainly focuses on mathematical and reasoning-intensive benchmarks. Although these settings provide a controlled testbed for studying test-time self-improvement, extending the evaluation to a broader range of task families would further strengthen the generality of the conclusions.
(2) To ensure fair comparison across methods, we adopt a largely unified test-time adaptation setup in our experiments. Exploring a wider design space of update frequency, synthesis budget, and iteration schedules may yield additional insights into the behavior of TTSR under different resource conditions.
(3) While our experiments include several representative models, the current study is not intended to be an exhaustive comparison across all model families, parameter scales, and alignment settings. A broader model coverage would provide a more complete picture of how TTSR behaves under different model configurations.
In future work, we aim to broaden the evaluation of TTSR beyond mathematical and reasoning-intensive benchmarks to more diverse domains and task settings. We are also interested in studying richer adaptation schedules and broader model coverage, which may provide a more comprehensive understanding of how TTSR scales across different resource budgets and model families.

\section*{Acknowledgments}

This work was supported by the Beijing Natural Science Foundation (Grant No. QY24214) and Brain Science and Brain-like Intelligence Technology-National Science and Technology Major Project (2021ZD0200600, 2021ZD0200601). We sincerely thank the anonymous reviewers and the Area Chairs for their insightful comments and constructive suggestions, which have greatly helped us improve this paper.

\bibliography{reference}

\appendix
\onecolumn

\section{Background: GRPO and TTT}
\label{app:preliminary}

This section briefly reviews the optimization and adaptation techniques underlying our framework, including Group Relative Policy Optimization (GRPO) ~\citep{shao2024deepseekmath} for reward-based learning and Test-Time Training (TTT)~\cite{zuo2026ttrl} for label-free model adaptation at inference time.

\paragraph{Group Relative Policy Optimization (GRPO).}
Group Relative Policy Optimization (GRPO)~\citep{shao2024deepseekmath} is a policy-gradient algorithm for large language models that removes the need for an explicit value function. Instead of estimating advantages via a learned critic, GRPO derives relative advantages by normalizing outcome-level rewards across a group of sampled responses, which enables stable optimization under sparse or binary supervision.

We model the language model as a stochastic policy $\pi_\theta$ that generates a response $y=(y_1,\dots,y_{|y|})$ conditioned on an input question $q$. For each question $q$, a behavior policy $\pi_{\theta_{\mathrm{old}}}$ samples a group of $G$ candidate responses $\{y_i\}_{i=1}^G$, each receiving a sequence-level reward $R_i$ from a verifier.

To derive learning signals without a value function, GRPO normalizes rewards within the sampled group and defines a relative advantage for each response:
\begin{equation}
\tilde{A}_i
=
\frac{R_i - \mathrm{mean}(\mathbf{R})}{\mathrm{std}(\mathbf{R}) + \delta},
\qquad
\mathbf{R} = \{R_1,\dots,R_G\},
\end{equation}
where $\delta$ is a small constant for numerical stability. The same advantage $\tilde{A}_i$ is shared across all token positions within response $y_i$.

GRPO updates the policy parameters by maximizing a PPO-style clipped surrogate objective with an explicit KL regularization term:
\begin{equation}
\begin{aligned}
\mathcal{J}&_{\mathrm{GRPO}}(\theta)
= \mathbb{E}_{q \sim \mathcal{D}_q,\,
\{y_i\}_{i=1}^{G} \sim \pi_{\theta_{\mathrm{old}}}}
\\
&\Biggl[
\frac{1}{G}\sum_{i=1}^{G}\frac{1}{|y_i|}
\sum_{t=1}^{|y_i|}
\Bigl(
\min\!\bigl(
r_{i,t}(\theta)\,\tilde{A}_i,\,
\mathrm{clip}(r_{i,t}(\theta),\, 1-\epsilon,\, 1+\epsilon)\,\tilde{A}_i
\bigr)
- \beta\,\mathbb{D}_{\mathrm{KL}}\!\left(
\pi_\theta \parallel \pi_{\theta_{\mathrm{old}}}
\right)
\Bigr)
\Biggr],
\end{aligned}
\label{eq:grpo_final}
\end{equation}

where
\begin{equation}
r_{i,t}(\theta)
=
\frac{\pi_{\theta}(y_{i,t} \mid q, y_{i,<t})}
     {\pi_{\theta_{\mathrm{old}}}(y_{i,t} \mid q, y_{i,<t})},
\qquad
\tilde{A}_i
=
\frac{R_i - \mathrm{mean}(\{R_i\}_{i=1}^{G})}
     {\mathrm{std}(\{R_i\}_{i=1}^{G}) + \delta}.
\label{eq:advantage_calculation}
\end{equation}

Here, $\epsilon$ controls the clipping range, and $\beta$ balances the KL regularization term, which constrains policy updates to remain close to the behavior policy and improves optimization stability.

\paragraph{Test-Time Training (TTT).}
Test-Time Training enables a pretrained policy to adapt its parameters during inference using only unlabeled test inputs. Let $\mathcal{X}_{\text{test}}=\{x^{(t)}\}$ denote the set of test queries.

Starting from an initial policy $\pi_{\theta_\text{old}}$, TTT seeks adapted parameters by optimizing a self-supervised objective:
\begin{equation}
\theta^{\dagger} =
\arg\max_{\theta}\;
\mathbb{E}_{y \sim \pi_\theta(\cdot \mid x^{(t)})}
\big[
\mathcal{R}_{\mathrm{TTT}}(y, \tilde{y})
\big],
\end{equation}
where $\tilde{y}$ denotes a pseudo-target constructed from the model's own predictions, typically obtained via self-consistency or consensus-based aggregation. The reward function $\mathcal{R}_{\mathrm{TTT}}$ measures agreement between a candidate response and the pseudo-target, providing a learning signal for inference-time adaptation.

\twocolumn
\section{Experiment Details}
\label{app:experiment_details}
\subsection{Benchmarks}
\label{app:benchmarks}

\paragraph{Mathematical Reasoning Benchmarks.}
\begin{itemize}
    \item \textbf{AMC23}~\cite{zeng2025simplerl}: Competition-style problems that require precise reasoning and careful case analysis.
    \item \textbf{MATH-500}~\cite{MATH-500}: A diverse set of mathematical topics (algebra, geometry, number theory, and combinatorics) used to evaluate structured mathematical reasoning.
    \item \textbf{Minerva}~\cite{Minerva} and \textbf{OlympiadBench}~\cite{OlympiadBench}: Challenging mathematical reasoning benchmarks covering quantitative STEM problems and olympiad-level competition problems, respectively, both requiring multi-step derivations and precise symbolic reasoning.
    \item \textbf{AIME 2024} and \textbf{AIME 2025}~\cite{li2025limr}: Highly challenging settings where a single reasoning error can invalidate the entire solution.
\end{itemize}

\paragraph{General Reasoning Benchmarks.}
\begin{itemize}
    \item \textbf{GPQA-Diamond}: Expert-level scientific question answering, requiring deep reasoning over specialized knowledge.
    \item \textbf{MMLU-Pro}: Broad multi-domain reasoning evaluation across a wide range of subjects.
\end{itemize}

\paragraph{Evaluation Metrics.}
\begin{itemize}
    \item \textbf{Mean@32} (AIME 2024/2025): Accuracy under multiple sampled reasoning trajectories.
    \item \textbf{Greedy Decoding (Pass@1)} (others): Greedy decoding accuracy.
\end{itemize}

\subsection{Training Setting}
\label{app:training_setting}

All models and benchmarks share the configuration below unless otherwise noted.
At each iteration, the Student is updated on both the original test questions and the Teacher-synthesized variants from the previous iteration.

\begin{table*}[t]
\centering
\scriptsize
\setlength{\tabcolsep}{2.5pt}
\renewcommand{\arraystretch}{1.1}
\resizebox{\textwidth}{!}{
\begin{tabular}{lccccccc}
\toprule
\textbf{Algorithm}
& \textbf{Batch Size}
& \textbf{Student Rollout $G$}
& \textbf{Teacher Variants $M$}
& \textbf{Iterations $T$}
& \textbf{KL Coef}
& \textbf{Learning Rate}
& \textbf{Max Len} \\
\midrule
GRPO
& 16
& 16
& 2
& 10
& 0.001
& $3\mathrm{e}{-7}$
& 4096 \\
\bottomrule
\end{tabular}
}
\caption{Important Training Parameters for Test-Time Adaptation.}
\label{tab:training_params}
\end{table*}
\paragraph{Similarity Penalty Configuration.}
To discourage redundant or near-duplicate question generation during reflection-guided synthesis, we incorporate a similarity-based penalty into the Teacher's reward.
Pairwise similarities are computed between synthesized variant questions and the original test question, as well as among the synthesized questions within the same batch.
The similarity threshold is fixed to $\tau = 0.75$, and the penalty coefficient is set to $\lambda = 1.0$ in all experiments.
These values are kept constant across models and benchmarks.

\section{Method Details}
\onecolumn
\subsection{Algorithmic Summary}
\label{app:algorithm}

Algorithm~\ref{alg:ttsr} summarizes the full TTSR procedure, including Student adaptation, Teacher reflection-guided synthesis, cross-iteration memory updates, and a rollout-reuse acceleration strategy that pre-evaluates synthesized variants and reuses these rollouts in the next iteration.

\begin{algorithm}[t]
\small
\begin{algorithmic}[1]
\REQUIRE Test set $\mathcal{X}_{\text{test}}$, initial shared policy $\pi_{\theta_0}$, iterations $T$, Student rollout count $G$, Teacher rollout batch size $M$, memory budget $K$, stale threshold $\Delta$, strategy-note size $N$
\STATE Initialize weakness memory $\mathcal{M}_0 \leftarrow \emptyset$, variant set $\mathcal{X}^{(0)}_{\text{var}} \leftarrow \emptyset$, cached rollout buffer $\mathcal{R}^{(0)}_{\text{var}} \leftarrow \emptyset$
\FOR{$t = 1$ to $T$}
    \STATE Compile strategy note $h_t \leftarrow \mathrm{Compile}(\mathcal{M}_{t-1}^{\mathrm{top}\text{-}N})$
    \STATE Construct mixed training set $\mathcal{D}_t \leftarrow \mathcal{X}_{\text{test}} \cup \mathcal{X}^{(t-1)}_{\text{var}}$
    \STATE Initialize failed-instance buffer $\mathcal{F}^{(t)} \leftarrow \emptyset$
    \FORALL{$x \in \mathcal{D}_t$}
        \IF{$x \in \mathcal{X}^{(t-1)}_{\text{var}}$ \textbf{and} cached rollout exists in $\mathcal{R}^{(t-1)}_{\text{var}}$}
            \STATE Load cached trajectories $\{y_i\}_{i=1}^G$, consensus answer $\hat{a}(x)$, and rewards $\{R_S(y_i \mid x)\}_{i=1}^G$
        \ELSE
            \STATE Sample Student trajectories $\{y_i\}_{i=1}^G \sim \pi_{\theta_{t-1}}(\cdot \mid h_t, x)$
            \STATE Compute consensus answer $\hat{a}(x)=\mathrm{Maj}(\{a(y_i)\}_{i=1}^{G})$ and rewards $R_S(y_i \mid x)=\mathbb{I}[a(y_i)=\hat{a}(x)]$
        \ENDIF
        \STATE Estimate pseudo-correctness score $s_t(x) \leftarrow \frac{1}{G}\sum_{i=1}^{G} R_S(y_i \mid x)$
        \STATE Add failed tuples $(x, y_i, \hat{a}(x))$ with $R_S(y_i \mid x)=0$ to $\mathcal{F}^{(t)}$
    \ENDFOR
    \STATE Apply Student-role GRPO to the shared policy using $R_S$, obtaining the intermediate policy $\pi_{\theta_t^{S}}$
    \STATE With the same shared policy $\pi_{\theta_t^{S}}$, reflect on a sampled subset of $\mathcal{F}^{(t)}$ to extract weakness descriptors $\mathcal{W}^{(t)}$
    \STATE Update memory:
    \[
    \mathcal{M}_t \leftarrow \mathrm{Evict}\!\Big(\mathrm{Resolve}\!\big(\mathrm{Merge}(\mathcal{M}_{t-1}, \mathcal{W}^{(t)}), \Delta\big), K\Big)
    \]
    \STATE Build synthesis prompt from sampled failures and persistent entries in $\mathcal{M}_t$
    \STATE Initialize next-round rollout cache $\mathcal{R}^{(t)}_{\text{var}} \leftarrow \emptyset$
    \STATE Sample one Teacher rollout batch $\mathcal{B}^{(t)}_T=\{x'_j\}_{j=1}^M \sim \pi_{\theta_t^{S}}(\cdot \mid \mathcal{F}^{(t)}, \mathcal{M}_t)$
    \STATE Batch-pre-evaluate $\mathcal{B}^{(t)}_T$ with Student rollouts $\{y'_{j,i}\}_{j=1,i=1}^{M,G} \sim \pi_{\theta_t^{S}}(\cdot \mid h_t, x'_j)$
    \STATE Batch-compute $\hat{a}(x'_j)$, $R_S(y'_{j,i}\mid x'_j)$, and $s_t(x'_j)$ for all $x'_j \in \mathcal{B}^{(t)}_T$
    \STATE Compute batch rewards $\mathbf{R}^{(t)}_T=\{\max(0,4s_t(x'_j)(1-s_t(x'_j))-\lambda P_{\mathrm{sim}}(x'_j,x))\}_{j=1}^M$
    \STATE Apply Teacher-role GRPO to the same shared policy on $\mathcal{B}^{(t)}_T$ using $\mathbf{R}^{(t)}_T$, obtaining $\pi_{\theta_t}$
    \STATE Cache the batch evaluation results in $\mathcal{R}^{(t)}_{\text{var}}$ for reuse in iteration $t+1$
    \STATE Set $\mathcal{X}^{(t)}_{\mathrm{var}} \leftarrow \mathcal{B}^{(t)}_T$
\ENDFOR
\RETURN Final adapted policy $\pi_{\theta_T}$, weakness memory $\mathcal{M}_T$, rollout cache $\mathcal{R}^{(T)}_{\text{var}}$, and strategy note compiled from $\mathcal{M}_T$
\end{algorithmic}
\caption{TTSR: Test-Time Self-Evolving via Reflection}
\label{alg:ttsr}
\end{algorithm}

\subsection{Similarity Metric Definition}
\label{app:similarity_metric}

To quantify the similarity between two questions in the similarity penalty, we adopt a sequence-based similarity measure that captures contiguous overlapping spans between two token sequences.
This metric is designed to penalize near-duplicate generations while remaining robust to minor surface variations.

\paragraph{Sequence Representation.}
Given two questions represented as token sequences $S_1$ and $S_2$, we define their total length as
\begin{equation}
T = |S_1| + |S_2|,
\end{equation}
where $|\cdot|$ denotes the number of tokens in the sequence.

\paragraph{Matching Blocks.}
We identify a set of non-overlapping matching blocks between $S_1$ and $S_2$.
Each matching block is defined as a contiguous subsequence that appears in both sequences.
The matching process first finds the longest contiguous matching subsequence and then recursively searches for additional matches in the remaining left and right segments.
Let the resulting set of matching blocks be represented as
\[
\mathcal{B} = \{(i, j, n)\},
\]
where $(i, j, n)$ denotes a matching block of length $n$ starting at position $i$ in $S_1$ and position $j$ in $S_2$.
The total number of matched tokens is then defined as
\begin{equation}
C = \sum_{(i,j,n) \in \mathcal{B}} n.
\end{equation}

\paragraph{Similarity Ratio.}
Using the above definitions, we compute the normalized similarity ratio between $S_1$ and $S_2$ as
\begin{equation}
\mathrm{sim}(S_1, S_2)
=
\frac{2C}{T},
\end{equation}
which takes values in the range $[0,1]$.
A higher similarity ratio indicates greater overlap between the two sequences, while a lower value reflects more substantial differences in surface form. This similarity measure is used to compute the pairwise similarity scores required by the similarity penalty described in the main text (Section~\ref{method:spr}).
By focusing on contiguous matching spans rather than isolated token overlap, the metric effectively discourages trivial paraphrasing while allowing moderate overlap that preserves shared reasoning structure.

\subsection{Theoretical Analysis of the Difficulty Reward}
\label{app:difficulty_reward_theory}

For a synthesized variant $x'$, the Student samples
$G$ trajectories
\[
\{y_i\}_{i=1}^{G}
\sim
\pi_{\theta_t}(\cdot\mid h_t,x'),
\]
constructs the majority-vote pseudo-answer
\[
\hat a(x')
=
\operatorname{Maj}
\bigl(\{a(y_i)\}_{i=1}^{G}\bigr),
\]
and assigns
\[
R_S(y_i\mid x')
=
\mathbb{I}[a(y_i)=\hat a(x')].
\]
Recall that
\[
s_t(x')
=
\frac{1}{G}\sum_{i=1}^{G}R_S(y_i\mid x'),
\qquad
R_{\mathrm{diff}}(x')
=
4s_t(x')(1-s_t(x')).
\]

We emphasize that $s_t(x')$ measures agreement with the
self-generated pseudo-answer. We analyze $R_{\mathrm{diff}}$ as the amount of
relative supervision available under this pseudo-labeling rule.

\paragraph{Finite-sample interpretation.}

For a fixed $x'$, let
\[
R_i:=R_S(y_i\mid x'),
\qquad
s:=s_t(x').
\]
Because $R_i\in\{0,1\}$,
\begin{equation}
\frac{1}{G}\sum_{i=1}^{G}(R_i-s)^2
=
s(1-s).
\label{eq:reward_variance}
\end{equation}
Hence,
\begin{equation}
R_{\mathrm{diff}}(x')
=
4\,\widehat{\mathrm{Var}}(R).
\label{eq:diff_as_variance}
\end{equation}
This identity is deterministic and does not require the
sample-dependent rewards $\{R_i\}$ to be independent.

An equivalent pairwise interpretation follows immediately. Let
\[
k=\sum_{i=1}^{G}R_i=Gs
\]
be the number of trajectories agreeing with $\hat a(x')$.
The number of positive--negative trajectory pairs is
$k(G-k)$, and therefore
\begin{equation}
R_{\mathrm{diff}}(x')
=
\frac{4k(G-k)}{G^2}.
\label{eq:diff_pairwise}
\end{equation}
Thus, $R_{\mathrm{diff}}$ is a normalized measure of the
within-group relative supervision. It is zero when all trajectories
receive the same pseudo-reward and is maximized at
$s_t(x')=1/2$, where positive--negative comparisons are most abundant.

This interpretation is directly aligned with GRPO, whose update is
driven by reward differences within a rollout group. In particular,
the average magnitude of the unnormalized centered reward satisfies
\begin{equation}
\frac{1}{G}\sum_{i=1}^{G}|R_i-s|
=
2s(1-s)
=
\frac{1}{2}R_{\mathrm{diff}}(x').
\label{eq:centered_reward}
\end{equation}
Therefore, larger $R_{\mathrm{diff}}$ corresponds to stronger
available group-relative reward contrast.


\paragraph{Stability of the majority-vote pseudo-answer.}
For a fixed $x'$, let
\[
p_{t,j}(x')
=
\Pr(a(y)=a_j\mid h_t,x'),
\qquad
y\sim\pi_{\theta_t}(\cdot\mid h_t,x').
\]
Assume a unique population modal answer, with
$p_{t,1}(x')>p_{t,2}(x')\ge\cdots\ge p_{t,K}(x')$.
Define
\[
a_t^\star(x')=a_1,\qquad
p_t^\star(x')=p_{t,1}(x'),\qquad
\Delta_t(x')=p_{t,1}(x')-p_{t,2}(x').
\]

\paragraph{Proposition 1 (Majority-vote stability).}
For $G$ independently sampled Student trajectories,
\begin{equation}
\Pr\!\left(
\hat a(x')\neq a_t^\star(x')
\right)
\le
(K-1)
\exp\!\left(
-\frac{G\Delta_t(x')^2}{2}
\right).
\label{eq:majority_stability}
\end{equation}

\noindent\textbf{Proof.}
For any competing answer $a_j$, $j\ge2$, define
\[
Z_i^{(j)}
=
\mathbb{I}[a(y_i)=a_1]
-
\mathbb{I}[a(y_i)=a_j].
\]
Then $Z_i^{(j)}\in[-1,1]$ and
\[
\mathbb{E}[Z_i^{(j)}]
=
p_{t,1}(x')-p_{t,j}(x')
\ge
\Delta_t(x').
\]
If $a_j$ receives at least as many votes as $a_1$, then
$\frac{1}{G}\sum_i Z_i^{(j)}\le0$.
Hoeffding's inequality gives
\[
\Pr(\hat p_j\ge\hat p_1)
\le
\exp\!\left(
-\frac{G\Delta_t(x')^2}{2}
\right).
\]

Taking a union bound over all competing answers
$j=2,\ldots,K$ yields
\[
\Pr\!\left(
\hat a(x')\neq a_t^\star(x')
\right)
\le
(K-1)
\exp\!\left(
-\frac{G\Delta_t(x')^2}{2}
\right),
\]
which proves Eq.~\eqref{eq:majority_stability}.

\paragraph{Finite-sample approximation of the frontier score.}

On the event
$\hat a(x')=a_t^\star(x')$,
$s_t(x')$ is simply the empirical frequency of the population modal
answer. Combining Proposition~1 with Hoeffding's inequality therefore
gives, for any $\epsilon>0$,
\begin{equation}
\Pr\!\left(
|s_t(x')-p_t^\star(x')|
\ge\epsilon
\right)
\le
(K-1)e^{-G\Delta_t(x')^2/2}
+
2e^{-2G\epsilon^2}.
\label{eq:score_bound}
\end{equation}

Define the population counterpart
\begin{equation}
R_{\mathrm{diff}}^\star(x')
=
4p_t^\star(x')
\bigl(1-p_t^\star(x')\bigr).
\end{equation}
Since $f(u)=4u(1-u)$ is $4$-Lipschitz on $[0,1]$,
Eq.~\eqref{eq:score_bound} immediately implies
\begin{equation}
\Pr\!\left(
|R_{\mathrm{diff}}(x')
-
R_{\mathrm{diff}}^\star(x')|
\ge4\epsilon
\right)
\le
(K-1)e^{-G\Delta_t(x')^2/2}
+
2e^{-2G\epsilon^2}.
\label{eq:reward_bound}
\end{equation}

The two terms have complementary meanings. The first controls whether
finite-sample voting identifies the Student's dominant answer, while
the second controls estimation error once that answer is identified.
Thus, increasing $G$ improves both pseudo-target stability and
frontier-score estimation.

\paragraph{Implication for curriculum synthesis.}

$R_{\mathrm{diff}}(x')$ measures the relative learning signal induced
by the current self-generated supervision. At the realized-group level,
$R_{\mathrm{diff}}$ is exactly the normalized binary reward variance
and is proportional to the number of positive--negative trajectory
pairs. At the population level, when the modal margin
$\Delta_t(x')$ is sufficiently large, majority voting provides a
stable pseudo-target and the empirical score approaches
$R_{\mathrm{diff}}^\star(x')$ as $G$ increases.

We therefore interpret regions of high
$R_{\mathrm{diff}}$ as a \emph{self-supervised learning frontier}:
the Student exhibits substantial within-group reward contrast,
providing informative relative supervision for GRPO. Reflection-guided
synthesis aims to construct variants near this frontier.

\subsection{Prompt Templates}
\begin{table*}[t]
\centering
\small
\colorbox{gray!10}{
\renewcommand{\arraystretch}{1.15}
\begin{tabular}{p{0.93\textwidth}}
\toprule
\rowcolor{gray!10}
\texttt{[ROLE: TEACHER]} \\[0.5em]

\texttt{Your task is to extract a generalizable reasoning weakness from an original question and a failed or unstable reasoning trace produced by a student model.} \\

\texttt{You must strictly base your analysis on the given inputs. Do not introduce external knowledge, do not assume access to the correct answer, and do not judge final correctness.} \\

\texttt{[ORIGINAL\_QUESTION]} \\
\texttt{\{text\}} \\[0.3em]

\texttt{[FAILED\_REASONING\_TRACE]} \\
\texttt{\{text\}} \\[0.3em]

\texttt{[ANALYSIS GUIDELINES]} \\
\texttt{(1) Error Localization: Identify the first point where the reasoning becomes unreliable, incomplete, or invalid.} \\
\texttt{(2) Weakness Abstraction: Summarize the underlying reasoning weakness in one abstract sentence, without referring to specific values or variables.} \\
\texttt{(3) Trigger Conditions: Describe what problem structures or conditions are likely to trigger this weakness.} \\
\texttt{(4) Failure Signature: Describe typical reasoning patterns or behaviors when this weakness appears.} \\[0.5em]

\texttt{[OUTPUT REQUIREMENTS]} \\
\texttt{Output must be valid JSON and contain the following fields:} \\

\texttt{\{ } \\
\texttt{\ \ "reasoning\_weakness": "...",} \\
\texttt{\ \ "trigger\_conditions": ["..."],} \\
\texttt{\ \ "failure\_signature": ["..."],} \\
\texttt{\ \ "localization\_summary": "..."} \\
\texttt{\}} \\[0.5em]

\texttt{[EXAMPLE]} \\

\texttt{[ORIGINAL\_QUESTION]} \\
\texttt{Let x and y be real numbers such that x + y = 10 and x\textasciicircum2 + y\textasciicircum2 = 58. Find the maximum possible value of x.} \\

\texttt{[FAILED\_REASONING\_TRACE]} \\
\texttt{From x + y = 10, y = 10 - x. Substituting gives x\textasciicircum2 + (10 - x)\textasciicircum2 = 58. Solving yields x = 3 or x = 7. The reasoning directly selects 7 as the maximum.} \\

\texttt{[OUTPUT]} \\
\texttt{\{ } \\
\texttt{\ \ "reasoning\_weakness": "When multiple valid solutions exist, the reasoning directly selects} \\
\texttt{\ \ \ \ the numerically larger solution without semantic justification.",} \\
\texttt{\ \ "trigger\_conditions": ["Multiple algebraic solutions", "Extremal objectives"],} \\
\texttt{\ \ "failure\_signature": ["Selecting the largest root immediately", "Equating magnitude with optimality"],} \\
\texttt{\ \ "localization\_summary": "The reasoning deviates when it selects the larger algebraic solution solely based on its value."} \\
\texttt{\}} \\

\bottomrule
\end{tabular}}
\caption{Teacher prompt template used for reasoning weakness extraction in TTSR.}
\label{tab:reflector_full_prompt}
\end{table*}

\begin{table*}[tp]
\centering
\small
\renewcommand{\arraystretch}{1.15}
\colorbox{gray!10}{
\begin{tabular}{p{0.93\textwidth}}
\toprule
\rowcolor{gray!10}

\texttt{[ROLE: TEACHER]} \\[0.5em]

\texttt{Your task is to generate one new training question based on the original question, a failed reasoning trace, and an extracted reasoning weakness.} \\

\texttt{The goal is not to make the question arbitrarily harder, but to produce a question that is targeted, learnable, and close to the student's capability frontier.} \\[0.5em]

\texttt{You must ensure that:} \\
\texttt{(1) The new question preserves the core reasoning structure of the original question.} \\
\texttt{(2) The new question is more likely to trigger the given reasoning weakness.} \\
\texttt{(3) The new question is self-contained, solvable, and unambiguous.} \\
\texttt{(4) The new question is not a superficial paraphrase or simple numerical substitution.} \\[0.5em]

\texttt{[INPUT]} \\

\texttt{[ORIGINAL\_QUESTION]} \\
\texttt{\{text\}} \\[0.3em]

\texttt{[FAILED\_REASONING\_TRACE]} \\
\texttt{\{text\}} \\[0.3em]

\texttt{[WEAKNESS\_JSON]} \\
\texttt{\{reasoning\_weakness, trigger\_conditions, failure\_signature\}} \\[0.5em]

\texttt{[STEP-BY-STEP INSTRUCTIONS]} \\

\texttt{Step 1: Anchor Structure.} \\
\texttt{Summarize the core reasoning structure of the original question, including key concepts, variable relationships, constraints, and the main reasoning steps required for a correct solution.} \\
\texttt{Use structural language rather than restating the original question.} \\[0.5em]

\texttt{Step 2: Error-Hitting Strategy.} \\
\texttt{Explicitly state how the new question will be designed to more reliably trigger the given reasoning weakness, while remaining solvable and fair.} \\
\texttt{Explain what aspects will be modified, what shortcuts are targeted, and how superficial paraphrasing is avoided.} \\[0.5em]

\texttt{Step 3: Generate One New Question.} \\
\texttt{Generate a single, fully self-contained question that requires multi-step reasoning and is likely to trigger the weakness under common shortcuts.} \\
\texttt{Avoid changing only numbers, swapping story contexts, or adding irrelevant complexity.} \\[0.5em]

\texttt{Step 4: Hit Rationale.} \\
\texttt{Briefly explain why the generated question is likely to expose the reasoning weakness and how a correct solution would avoid the shortcut.} \\[0.5em]

\texttt{Step 5: Self-Test Filter.} \\
\texttt{Verify whether the question is likely to trigger the weakness, remains learnable, and is not a surface-level paraphrase.} \\
\texttt{If any check fails, regenerate the question.} \\[0.5em]

\bottomrule
\end{tabular}}
\caption{Teacher prompt template used for reflection-guided question synthesis in TTSR.}
\label{tab:challenger_full_prompt}
\end{table*}

\begin{table*}[tp]
\centering
\small
\renewcommand{\arraystretch}{1.15}
\colorbox{gray!10}{
\begin{tabular}{p{0.93\textwidth}}
\toprule
\rowcolor{gray!10}

\texttt{[OUTPUT REQUIREMENTS]} \\[0.4em]

\texttt{The output must be valid JSON. Do not include markdown or additional explanations.} \\
\texttt{All fields must be present and filled according to the schema below.} \\[0.6em]

\texttt{\{} \\

\texttt{\ \ "anchor\_structure": [} \\
\texttt{\ \ \ \ "...core structural element 1...",} \\
\texttt{\ \ \ \ "...core structural element 2..."} \\
\texttt{\ \ ],} \\[0.4em]

\texttt{\ \ "error\_hitting\_strategy": \{} \\
\texttt{\ \ \ \ "what\_to\_avoid": ["...simple variants that would not trigger the weakness..."],} \\
\texttt{\ \ \ \ "what\_to\_add": ["...structural modifications to increase weakness exposure..."],} \\
\texttt{\ \ \ \ "shortcut\_to\_block": ["...erroneous shortcut to be induced or blocked..."],} \\
\texttt{\ \ \ \ "fairness\_check": "...how solvability and unambiguity are ensured..."} \\
\texttt{\ \ \},} \\[0.4em]

\texttt{\ \ "generated\_question": "...fully self-contained question text...",} \\[0.4em]

\texttt{\ \ "hit\_rationale": [} \\
\texttt{\ \ \ \ "...why the question is likely to trigger the weakness...",} \\
\texttt{\ \ \ \ "...what a correct reasoning process would need to check..."} \\
\texttt{\ \ ],} \\[0.4em]

\texttt{\ \ "self\_test": \{} \\
\texttt{\ \ \ \ "likely\_to\_trigger\_weakness": "YES/NO + brief reason",} \\
\texttt{\ \ \ \ "learnable\_frontier": "YES/NO + brief reason",} \\
\texttt{\ \ \ \ "not\_surface\_paraphrase": "YES/NO + brief reason"} \\
\texttt{\ \ \}} \\

\texttt{\}} \\

\bottomrule
\end{tabular}}
\caption{Teacher prompt template used for reflection-guided question synthesis in TTSR (continued).}
\label{tab:challenger_output_schema}
\end{table*}

\twocolumn
\section{Case Study}
\label{app:case_study}

To qualitatively illustrate how the TTSR pipeline operates, we present two representative case studies from Qwen3-8B-Base trained on MATH500.

\paragraph{Full pipeline trace (Table~\ref{tab:case_study_pipeline}).}
Table~\ref{tab:case_study_pipeline} traces a single training instance through the complete TTSR pipeline.
The Student encounters a quartic equation and solves it via the standard substitution $t = x^2$, but recovers only the positive roots---a failure pattern commonly observed in early iterations.
The Teacher analyzes the failed trace and extracts a structured weakness descriptor pinpointing the incomplete back-substitution.
Based on this weakness, the Teacher synthesizes a variant question that algebraically reduces to the same equation but asks for the \emph{sum} of all solutions, making the back-substitution error directly observable in the final answer.

\paragraph{Strategy note effectiveness (Table~\ref{tab:case_study_note}).}
Table~\ref{tab:case_study_note} demonstrates how the accumulated weakness memory translates into actionable reasoning guidance.
After the back-substitution weakness is recorded and persists across multiple iterations, it is compiled into a strategy note prepended to the Student's prompt.
When the Student encounters a new even-power equation at a later iteration, the strategy note explicitly reminds it to check for negative branches---resulting in a correct solution that the earlier Student would have missed.
This before-and-after comparison highlights how TTSR's memory mechanism enables \emph{targeted, persistent} reasoning improvement, rather than relying solely on gradient updates.

\paragraph{Variant evolution across iterations (Table~\ref{tab:case_study_evolution}).}
Table~\ref{tab:case_study_evolution} tracks how the Teacher's synthesized variant questions evolve over training iterations for two representative MATH500 problems.
In early iterations (Iter~2), variants are surface-level modifications: changing coefficients or base/modulus while preserving the same solution technique.
As the weakness memory accumulates, the Teacher produces increasingly targeted variants:
at Iter~5, variants \emph{amplify} the identified weakness (e.g., requiring Newton's identity chains instead of single-step Vieta's);
at Iter~8, variants \emph{transfer} the weakness to a different mathematical domain (e.g., reciprocal-squared roots, or modular arithmetic with composite moduli);
by Iter~10, variants \emph{integrate} multiple weaknesses into a single problem requiring simultaneous application of several reasoning techniques.
This progression demonstrates that TTSR's reflection-guided synthesis produces a natural curriculum---from surface variation to deep structural targeting---without any explicit difficulty scheduling.

\definecolor{csblue}{HTML}{E3F2FD}
\definecolor{csred}{HTML}{FFEBEE}
\definecolor{csyellow}{HTML}{FFF8E1}
\definecolor{csgreen}{HTML}{E8F5E9}
\definecolor{cspurple}{HTML}{F3E5F5}

\begin{table*}[t]
\centering
\small
\renewcommand{\arraystretch}{1.2}
\begin{tabular}{@{}p{0.97\textwidth}@{}}
\toprule

\cellcolor{csblue}
\textbf{[Original Question]}~~\textit{(Source: MATH500, Iteration 3)} \\
\cellcolor{csblue}
Find the number of distinct real solutions to the equation $x^4 - 13x^2 + 36 = 0$. \\[4pt]

\cellcolor{csred}
\textbf{[Student's Failed Reasoning Trace]} \\
\cellcolor{csred}
\texttt{Let $t = x^2$. Then $t^2 - 13t + 36 = 0$.} \\
\cellcolor{csred}
\texttt{Using the quadratic formula: $t = \frac{13 \pm \sqrt{169 - 144}}{2} = \frac{13 \pm 5}{2}$.} \\
\cellcolor{csred}
\texttt{So $t = 9$ or $t = 4$.} \\
\cellcolor{csred}
\texttt{From $t = 9$: $x = 3$.~~From $t = 4$: $x = 2$.} \\
\cellcolor{csred}
\texttt{Therefore, there are \textcolor{red}{\textbf{2}} distinct real solutions.}~~{\small\textcolor{red}{($\times$~Correct answer: \textbf{4})}} \\[4pt]

\cellcolor{csyellow}
\textbf{[Teacher's Extracted Weakness]}  \\
\cellcolor{csyellow}
\texttt{\{}  \\
\cellcolor{csyellow}
\texttt{~~"reasoning\_weakness": "After solving via even-power substitution,}\\
\cellcolor{csyellow}
\texttt{~~~~only the principal (positive) root is recovered, omitting the negative branch.",}\\
\cellcolor{csyellow}
\texttt{~~"trigger\_conditions": ["Even-power variable substitution ($t = x^n$, $n$ even)",}\\
\cellcolor{csyellow}
\texttt{~~~~~~~"Requests for the count or aggregate of all real solutions"],}\\
\cellcolor{csyellow}
\texttt{~~"failure\_signature": ["Only non-negative roots reported",}\\
\cellcolor{csyellow}
\texttt{~~~~~~~"Solution count is exactly half the true count"],}\\
\cellcolor{csyellow}
\texttt{~~"localization\_summary": "Deviation occurs at back-substitution:}\\
\cellcolor{csyellow}
\texttt{~~~~$t = 9$ yields $x = 3$ only, instead of $x = \pm 3$."}\\
\cellcolor{csyellow}
\texttt{\}} \\[4pt]

\cellcolor{csgreen}
\textbf{[Teacher's Synthesized Variant Question]} \\
\cellcolor{csgreen}
Let $S$ be the set of all real numbers $x$ satisfying $(x^2 - 4)(x^2 - 9) = 0$. Compute the \textit{sum} of the elements of $S$. \\
\cellcolor{csgreen}
\textit{Hit rationale}: Expanding yields $x^4 - 13x^2 + 36 = 0$---the same substitution equation. However, asking for the \textbf{sum} (rather than the count) makes the error observable: the incorrect answer is $2 + 3 = 5$, while the correct answer accounting for $\pm$ branches is $(-3) + (-2) + 2 + 3 = 0$. \\

\bottomrule
\end{tabular}
\caption{
\textbf{Case study: full TTSR pipeline on a representative example.}
We trace one training instance through the complete pipeline---from the original question through the Student's failed reasoning, the Teacher's weakness extraction, and the synthesized variant question.
Color coding: \colorbox{csblue}{original question}, \colorbox{csred}{failed trace}, \colorbox{csyellow}{Teacher reflection}, \colorbox{csgreen}{variant question}.}
\label{tab:case_study_pipeline}
\end{table*}

\begin{table*}[t]
\centering
\small
\renewcommand{\arraystretch}{1.2}
\begin{tabular}{@{}p{0.97\textwidth}@{}}
\toprule

\cellcolor{csblue}
\textbf{[New Question]}~~\textit{(Encountered at Iteration 7)} \\
\cellcolor{csblue}
Find the sum of all real values of $x$ satisfying $x^6 = 64$.  \\[4pt]

\cellcolor{csred}
\textbf{[Student Reasoning — Without Strategy Note]}~~\textit{(Iteration 3 behavior)} \\
\cellcolor{csred}
\texttt{$x^6 = 64$. Taking the sixth root of both sides: $x = 64^{1/6} = 2$.} \\
\cellcolor{csred}
\texttt{The sum of all real values is \textcolor{red}{\textbf{2}}.}~~{\small\textcolor{red}{($\times$~Correct answer: \textbf{0})}} \\[4pt]

\cellcolor{cspurple}
\textbf{[Compiled Strategy Note]}~~\textit{(From weakness memory at Iteration 7)} \\
\cellcolor{cspurple}
\texttt{[REASONING STRATEGY NOTES]} \\
\cellcolor{cspurple}
\texttt{Based on recurring reasoning patterns, keep the following in mind:} \\
\cellcolor{cspurple}
\texttt{1. When solving equations via substitution $t = x^n$ with $n$ even, always} \\
\cellcolor{cspurple}
\texttt{~~~recover \textbf{both} $\pm \sqrt[n]{t}$ during back-substitution. Verify that} \\
\cellcolor{cspurple}
\texttt{~~~the solution count matches the degree of the original equation.} \\
\cellcolor{cspurple}
\texttt{2. When computing aggregates (sum, product) over all solutions,} \\
\cellcolor{cspurple}
\texttt{~~~enumerate every branch before combining.} \\[4pt]

\cellcolor{csgreen}
\textbf{[Student Reasoning — With Strategy Note]}~~\textit{(Iteration 7 behavior)} \\
\cellcolor{csgreen}
\texttt{$x^6 = 64$. Since the exponent $6$ is even, $x$ can be positive or negative.} \\
\cellcolor{csgreen}
\texttt{$|x|^6 = 64 \implies |x| = 2$, so $x = 2$ or $x = -2$.} \\
\cellcolor{csgreen}
\texttt{Verification: $2^6 = 64$~\checkmark;~~$(-2)^6 = 64$~\checkmark.} \\
\cellcolor{csgreen}
\texttt{The sum of all real values is $2 + (-2) = $ \textcolor{teal}{\textbf{0}}.}~~{\small\textcolor{teal}{(\checkmark~Correct)}} \\

\bottomrule
\end{tabular}
\caption{
\textbf{Case study: strategy note effectiveness on a subsequent problem.}
After the weakness from Table~\ref{tab:case_study_pipeline} is accumulated in the weakness memory, the compiled strategy note guides the Student to avoid the same error on a new problem.
Color coding: \colorbox{csred}{reasoning without note}, \colorbox{cspurple}{strategy note}, \colorbox{csgreen}{reasoning with note}.}
\label{tab:case_study_note}
\end{table*}

\definecolor{evoblue}{HTML}{1565C0}
\definecolor{evored}{HTML}{C62828}
\definecolor{evogreen}{HTML}{2E7D32}
\definecolor{evopurple}{HTML}{6A1B9A}

\begin{table*}[t]
\centering
\small
\renewcommand{\arraystretch}{1.25}
\setlength{\tabcolsep}{3pt}
\begin{tabular}{@{} >{\bfseries}l p{0.41\textwidth} p{0.41\textwidth} @{}}
\toprule
\textbf{Iteration}
& \textbf{Case 1: Symmetric Root Functions}
& \textbf{Case 2: Modular Exponentiation} \\
\midrule

\raggedright Test Question
&
Let $r_1, r_2, r_3$ be the roots of $x^3 - 7x + 6 = 0$.\newline
Find $r_1^2 + r_2^2 + r_3^2$.
&
Find the remainder when $2^{100}$ is divided by $7$.
\\[3pt]

\raggedright\scriptsize Extracted\newline Weakness
&
{\scriptsize\itshape
``Applies single-step Vieta's formula correctly but fails to chain identities for higher-order or non-standard symmetric functions of roots.''}
&
{\scriptsize\itshape
``Identifies cyclic patterns in modular arithmetic but fails to decompose nested exponents or composite moduli requiring multi-level analysis.''}
\\
\midrule

Iter 2
&
Find $r_1^2 + r_2^2 + r_3^2$ for the roots of
$x^3 \textcolor{evoblue}{- 3x^2 + x - 5} = 0$.
\newline
{\scriptsize\textcolor{evoblue}{[Surface: changed coefficients, same one-step Vieta]}}
&
Find the remainder when $\textcolor{evoblue}{3^{200}}$ is divided by~$\textcolor{evoblue}{13}$.
\newline
{\scriptsize\textcolor{evoblue}{[Surface: different base/modulus, same cycle technique]}}
\\[4pt]

Iter 5
&
For the roots $r_i$ of $x^3 - 4x^2 + 5x - 2 = 0$,
compute $\textcolor{evored}{r_1^3 + r_2^3 + r_3^3}$.
\newline
{\scriptsize\textcolor{evored}{[Weakness amplification: cubic power sum requires Newton's identity chain, not one-step Vieta]}}
&
What is the \textcolor{evored}{units digit} of $\textcolor{evored}{7^{7^7}}$?
\newline
{\scriptsize\textcolor{evored}{[Weakness amplification: nested exponent requires two-level cycle decomposition]}}
\\[4pt]

Iter 8
&
If the roots of $x^3 - 6x^2 + 11x - 6 = 0$ are $r_i$, find
$\textcolor{evogreen}{\dfrac{1}{r_1^2} + \dfrac{1}{r_2^2} + \dfrac{1}{r_3^2}}$.
\newline
{\scriptsize\textcolor{evogreen}{[Domain transfer: reciprocal-squared form shifts from polynomial to rational algebra]}}
&
Find the \textcolor{evogreen}{last two digits} of $3^{2025}$.
\newline
{\scriptsize\textcolor{evogreen}{[Domain transfer: mod~$100$ requires Euler's theorem or CRT decomposition]}}
\\[4pt]

Iter 10
&
The roots of $x^3 - x^2 - 4x + 3 = 0$ are $a,b,c$.\newline
Find $\textcolor{evopurple}{\displaystyle\sum_{\mathrm{cyc}} \frac{a^2}{b+c}}$.
\newline
{\scriptsize\textcolor{evopurple}{[Multi-weakness: fractional symmetric function combines identity chaining with constraint $b\!+\!c = S_1 - a$]}}
&
Determine the remainder when $\textcolor{evopurple}{2^{2^{2^3}}}$ is divided by~$\textcolor{evopurple}{1000}$.
\newline
{\scriptsize\textcolor{evopurple}{[Multi-weakness: tower exponent + 3-prime modulus combines nested cycle with CRT]}}
\\

\bottomrule
\end{tabular}
\caption{
\textbf{Evolution of reflection-guided variant questions across training iterations.}
Two representative cases from MATH500 are shown.
Unlike generic question synthesis, each variant is guided by the Teacher's \emph{extracted weakness}, resulting in progressively more targeted and structurally diverse problems.
\textcolor{evoblue}{Blue}: surface variation;
\textcolor{evored}{Red}: weakness amplification;
\textcolor{evogreen}{Green}: domain transfer;
\textcolor{evopurple}{Purple}: multi-weakness integration.
}
\label{tab:case_study_evolution}
\end{table*}

\section{Variant Quality Analysis}
\label{app:variant_quality}

\subsection{Frontier Score Distribution}

Table~\ref{tab:variant_quality} reports the pseudo-correctness score distribution of three question sources on Qwen3-8B-Base (AIME25): original test questions, variants synthesized without reflection (random structural variation), and TTSR variants with full reflection-guided synthesis.

\begin{table*}[t]
\centering
\small
\begin{tabular}{@{}lccc@{}}
\toprule
\textbf{Question Source} & \textbf{Mean $s_t$} & \textbf{Std} & \textbf{Frontier \%} \\
\midrule
Original questions      & 0.12 & 0.21 &  8.3 \\
w/o Reflection variants & 0.31 & 0.26 & 35.6 \\
TTSR variants           & 0.47 & 0.19 & 64.2 \\
\bottomrule
\end{tabular}
\caption{Variant quality on Qwen3-8B-Base (AIME25). Frontier ratio: fraction of questions with $s_t \in [0.2, 0.8]$.}
\label{tab:variant_quality}
\end{table*}

TTSR variants exhibit a substantially higher frontier ratio ($64.2\%$) than both original questions ($8.3\%$) and reflection-free variants ($35.6\%$), confirming that weakness-aware diagnosis effectively concentrates synthesized questions near the Student's capability boundary.

\subsection{Variant Structural Quality (GPT-5.2 Evaluation)}

Beyond frontier alignment, a key design goal is that TTSR variants target the \emph{specific weakness} identified by the Teacher's reflection, rather than being superficial modifications (e.g., changing numbers or swapping variable names).
To evaluate this, we use GPT-5.2 (2025-12-11) as an external judge on $n\!=\!100$ MATH500 and $n\!=\!50$ AIME25 variant instances.
Each instance is scored along two binary dimensions:

\begin{itemize}
    \item \textbf{Weakness Targeting (WT)}: Does the variant specifically test the reasoning weakness identified in the Teacher's reflection descriptor?
    \item \textbf{Structural Novelty (SN)}: Does the variant involve meaningful structural changes (e.g., domain transfer, constraint modification, multi-step integration) beyond simple number substitution or surface paraphrasing?
\end{itemize}

\begin{table*}[t]
\centering
\small
\begin{tabular}{@{}lcccc@{}}
\toprule
\textbf{Method} & \textbf{Benchmark} & \textbf{WT (\%)} & \textbf{SN (\%)} & \textbf{WT$\wedge$SN (\%)} \\
\midrule
w/o Reflection & MATH500 & ---   & 41.0 & --- \\
w/o Reflection & AIME25  & ---   & 22.0 & --- \\
\midrule
TTSR           & MATH500 & 83.0  & 78.0 & 71.0 \\
TTSR           & AIME25  & 48.0  & 46.0 & 38.0 \\
\bottomrule
\end{tabular}
\caption{GPT-5.2 (2025-12-11) structural evaluation of synthesized variants on Qwen3-8B-Base. WT: weakness targeting rate; SN: structural novelty rate; WT$\wedge$SN: both criteria satisfied jointly.}
\label{tab:variant_structural}
\end{table*}

Table~\ref{tab:variant_structural} shows that without reflection, the majority of variants ($59\%$ on MATH500, $78\%$ on AIME25) are trivial modifications, such as changing coefficients, swapping story contexts, or adding irrelevant complexity.
TTSR's reflection-guided synthesis raises the structural novelty rate to $78\%$ and $46\%$, while targeting the identified weakness in $83\%$ and $48\%$ of cases.
The joint rate (WT$\wedge$SN) is $71\%$ on MATH500 but only $38\%$ on AIME25, showing that fewer than half of the AIME25 variants satisfy both criteria.
The substantially lower AIME25 scores are consistent with the reflection quality evaluation in Appendix~\ref{app:reflection_eval}: competition-level problems produce longer reasoning chains where weakness localization is harder.

\subsection{Pseudo-Label Accuracy (GPT-5.2 Verification)}

\begin{figure*}[t]
    \centering
    \includegraphics[width=0.75\linewidth]{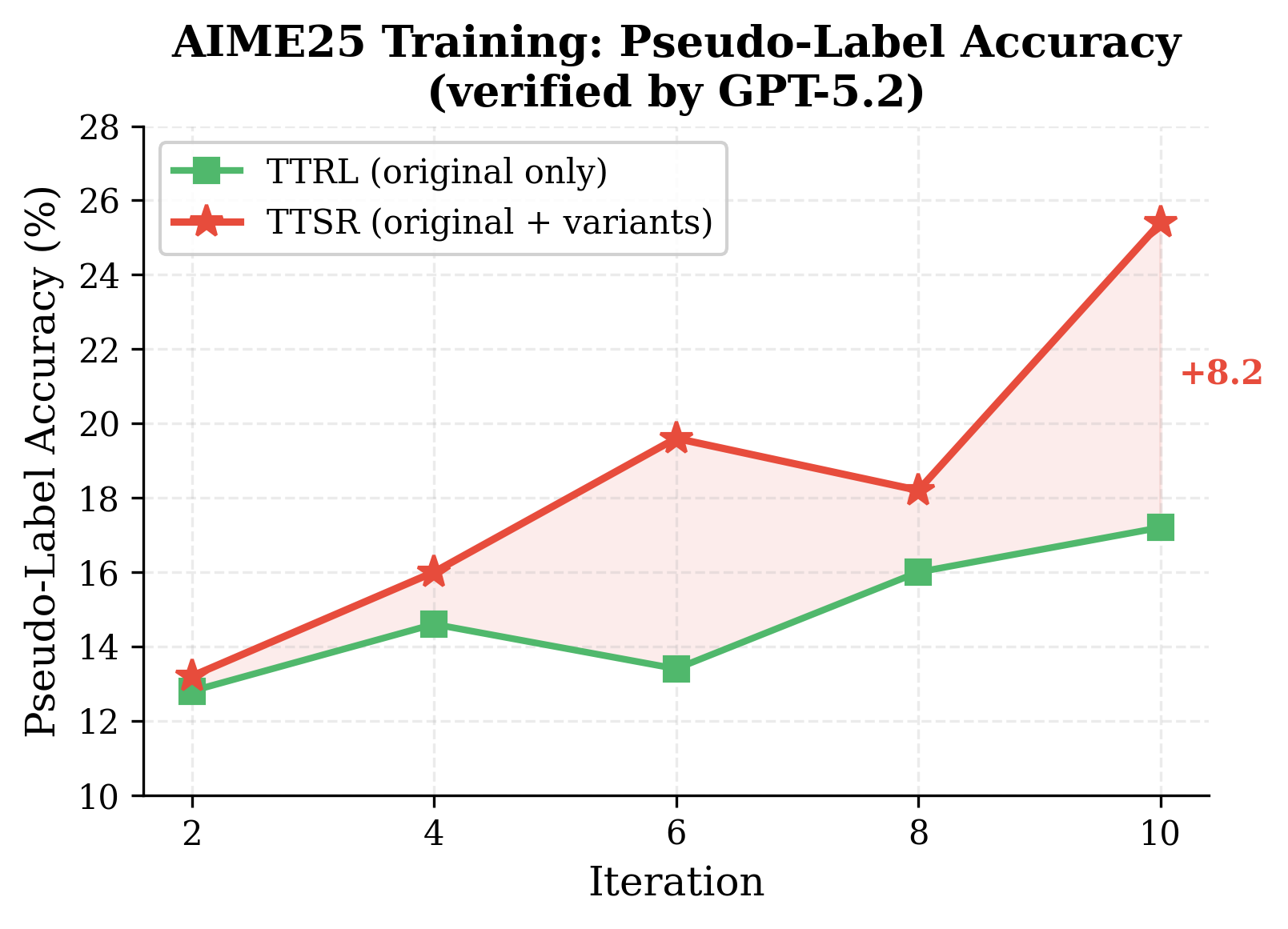}
    \caption{Majority-vote accuracy during AIME25 training (Qwen3-8B-Base). Original questions are verified by ground truth; variant questions are verified by GPT-5.2 (2025-12-11).}
    \label{fig:pseudo_label}
\end{figure*}

We track the majority-vote accuracy of the Student's training batch across iterations on AIME25 (Qwen3-8B-Base).
For original test questions, correctness is determined by the ground-truth answer; for synthesized variants (which have no canonical answer), we use GPT-5.2 as an oracle judge.
Figure~\ref{fig:pseudo_label} reports the results.

Both methods start near $13\%$ at iteration~2, reflecting the extreme difficulty of AIME25 (base accuracy $9.8\%$).
TTRL, which trains only on the original questions, improves slowly and with fluctuation, reaching $17.2\%$ at convergence.
TTSR introduces easier, targeted variant questions that provide more learnable samples and simultaneously raise the overall majority-vote accuracy to $25.4\%$ ($+8.2$ over TTRL).
The gap widens primarily in later iterations as training progresses and the Teacher synthesizes increasingly well-calibrated variants.

\section{Out-of-Distribution Performance}
\label{app:ood}

\begin{figure*}[t]
    \centering
    \includegraphics[width=\linewidth]{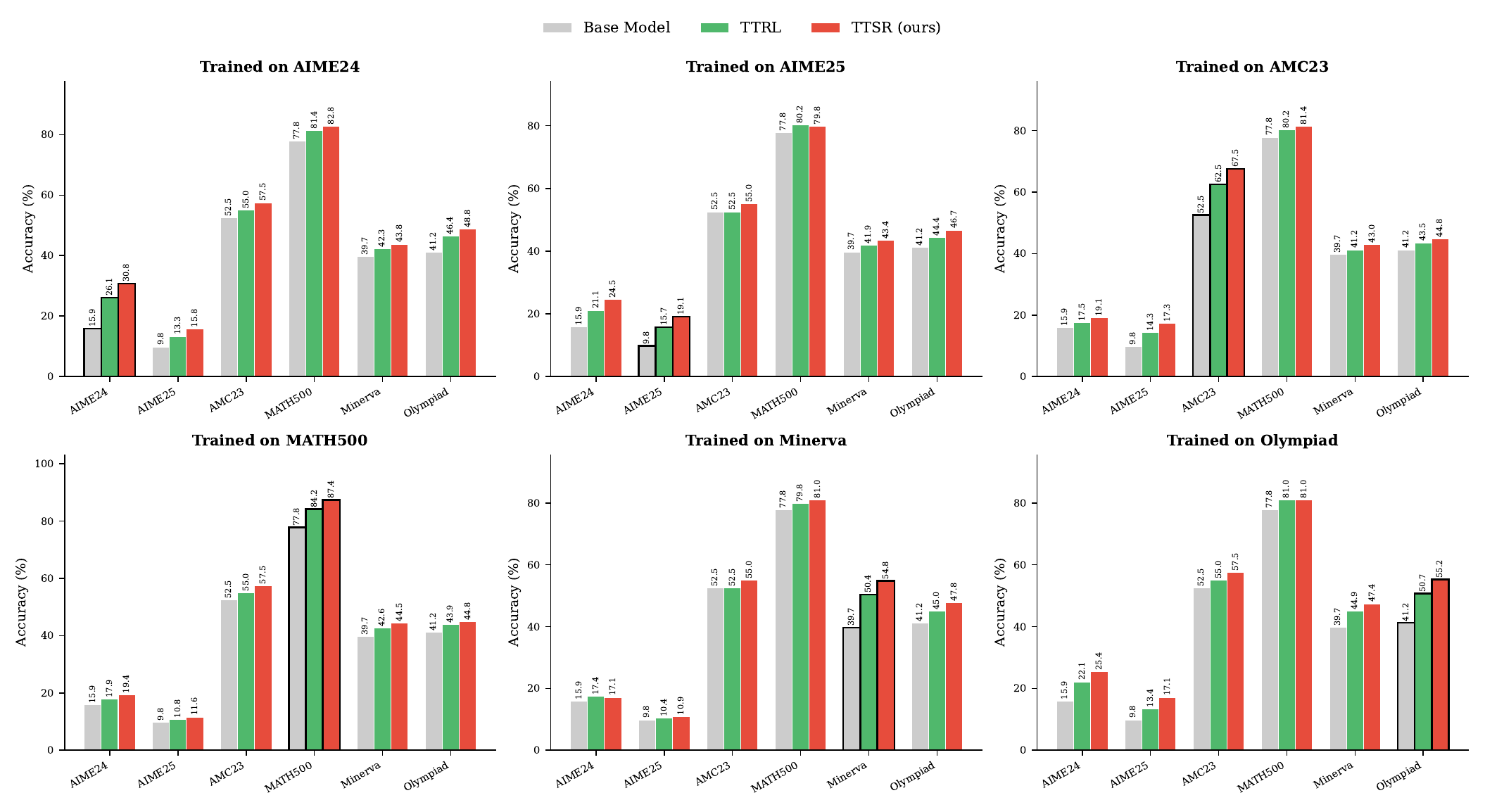}
    \caption{
    \textbf{Out-of-distribution performance comparison on Qwen3-8B-Base.}
    Each subplot shows the accuracy when the model is trained on a single source benchmark (title) and evaluated on all six mathematical reasoning benchmarks.
    In-domain bars (where train$=$eval) are highlighted with black borders.
    TTSR generally outperforms TTRL across most OOD evaluations.
    }
    \label{fig:ood_performance}
\end{figure*}

To assess the transferability of test-time training across different mathematical benchmarks, we train on each of the six source benchmarks independently and evaluate on all six benchmarks.
Figure~\ref{fig:ood_performance} reports the results on Qwen3-8B-Base.

\textbf{Cross-benchmark transfer.}
TTSR outperforms TTRL in the majority of OOD evaluations, with gains typically ranging from $1$--$3$ points.
For instance, when trained on Olympiad and evaluated on AIME24, TTSR reaches $25.4$ compared to TTRL's $22.1$ ($+3.3$), whereas the Base model achieves only $15.9$.
Notably, a small number of OOD transfer pairs exhibit comparable or slightly reversed results (e.g., AIME25$\rightarrow$MATH500: TTRL $80.2$ vs.\ TTSR $79.8$), which is expected given the stochastic nature of cross-benchmark generalization and the relatively small OOD gaps.

\textbf{Role of source difficulty.}
Training on harder benchmarks (AIME24, AIME25, Olympiad) generally produces stronger OOD transfer than training on easier ones (AMC23, MATH500).
For example, models trained on AIME24 achieve higher OOD scores on Olympiad ($48.8$ vs.\ $44.8$ from AMC23 training), suggesting that harder source problems induce more generalizable reasoning refinements.
TTSR amplifies this effect through reflection-guided synthesis: by targeting the model's specific weaknesses, the generated variant questions capture transferable reasoning patterns rather than benchmark-specific shortcuts.

\section{Additional Training Dynamics}
\label{app:training_dynamics}

\begin{figure*}[t]
    \centering
    \includegraphics[width=\linewidth]{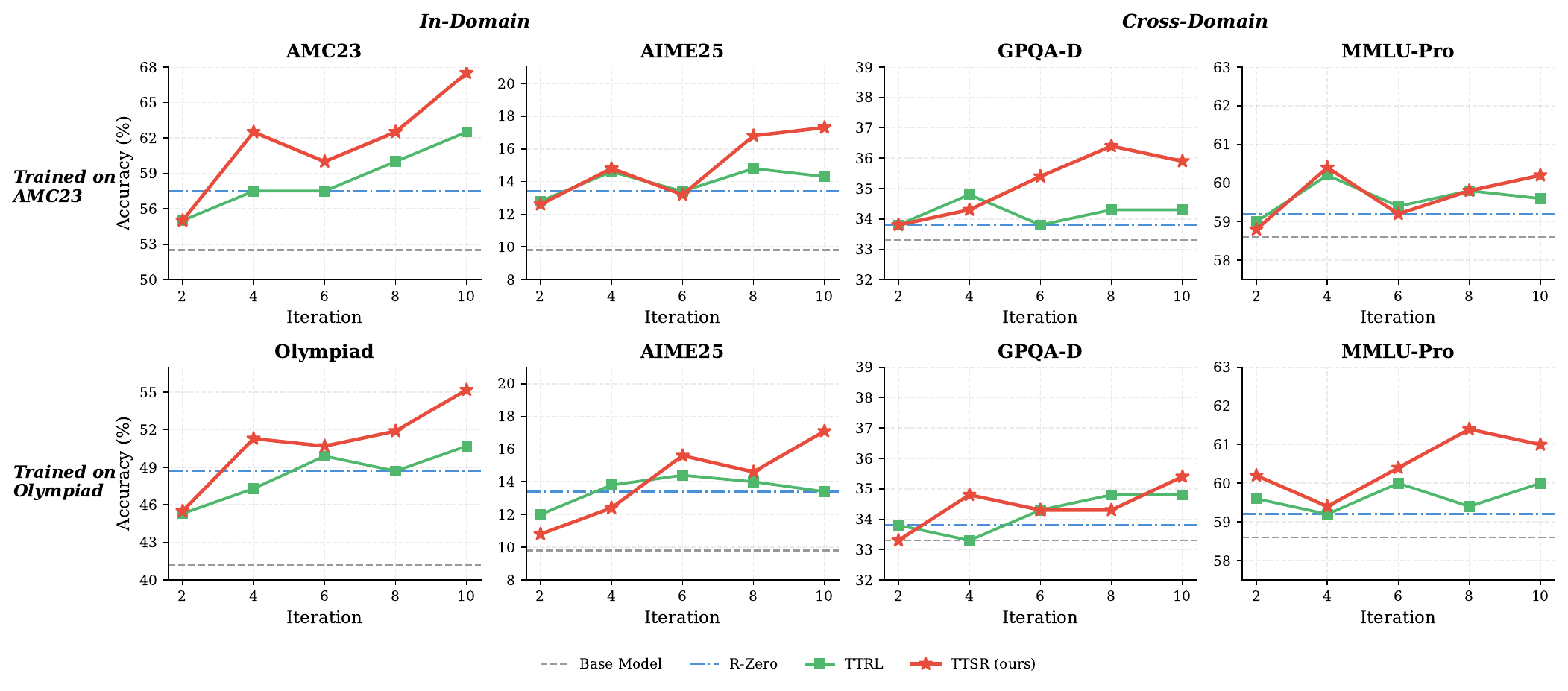}
    \caption{
    \textbf{Training dynamics on additional mathematical benchmarks (Qwen3-8B-Base).}
    Each row corresponds to a separate experiment where TTSR is applied with the indicated primary training benchmark.
    Left two panels show \emph{in-domain} accuracy (primary benchmark and AIME25);
    right two panels show \emph{cross-domain} accuracy on held-out general reasoning benchmarks (GPQA-D, MMLU-Pro).
    Dashed and dash-dotted horizontal lines denote the Base Model and R-Zero baselines, respectively.
    }
    \label{fig:training_dynamics_appendix}
\end{figure*}

To complement the MATH500-based training dynamics presented in Figure~\ref{fig:training_dynamics} (main text), we report additional training dynamics when TTSR is applied with AMC23 and Olympiad as primary training benchmarks.

\textbf{AMC23 (top row).}
On the in-domain AMC23 benchmark, TTSR matches TTRL at iteration~2, overtakes it by iteration~4, and reaches $67.5$ at convergence, a $+5.0$ margin over TTRL ($62.5$).
A mid-training dip at iteration~6 mirrors the pattern observed on MATH500 in the main text.
Cross-math transfer to AIME25 is moderate ($17.3$), reflecting the distributional gap between competition-format AMC problems and the AIME format.

\textbf{Olympiad (bottom row).}
Training on Olympiad produces a qualitatively different trajectory: TTSR jumps ahead by iteration~4 ($51.3$ vs.\ $47.3$) and, after a brief dip at iteration~6, resumes climbing to reach $55.2$ ($+4.5$ over TTRL at $50.7$) at convergence.
TTRL exhibits a mid-training dip at iteration~8 ($49.9 \!\to\! 48.7$) before recovering, whereas TTSR's gap over TTRL widens steadily from iteration~6 onward.
Cross-math transfer to AIME25 reaches $17.1$, and both rows confirm that cross-domain gains on GPQA-D and MMLU-Pro remain positive but modest.

\section{Self-Consistency Comparison}
\label{app:sc_comparison}

To disentangle the effect of inference-time scaling from test-time training, we compare against a \textbf{Self-Consistency} (SC)~\cite{wang2022self} baseline that applies majority voting over $G\!=\!16$ sampled reasoning trajectories per question \emph{without any parameter updates}.
Table~\ref{tab:sc_comparison} reports results for all three base models across eight benchmarks.

\begin{table*}[t]
\centering
\small
\renewcommand{\arraystretch}{1.05}
\setlength{\tabcolsep}{3pt}
\begin{tabular}{@{}lccccccccc@{}}
\toprule
\textbf{Method} & \textbf{AMC23} & \textbf{MATH500} & \textbf{Minerva} & \textbf{Olympiad} & \textbf{AIME24} & \textbf{AIME25} & \textbf{GPQA-D} & \textbf{MMLU-Pro} & \textbf{$\Delta$} \\
\midrule
\multicolumn{10}{@{}l}{\cellcolor{cyan!6}\textit{Qwen3-4B-Base}} \\
Base Model     & 45.0 & 72.0 & 32.4 & 40.2 & 12.4 & 5.8 & 25.8 & 52.0 & -- \\
SC & 52.5 & 77.0 & 37.9 & 43.2 & 16.8 & 8.8 & 28.3 & 54.5 & +4.2 \\
R-Zero         & 55.0 & 76.8 & 40.8 & 44.1 & 18.2 & 9.1 & 28.3 & 55.0 & +5.2 \\
TTRL           & 55.0 & 79.0 & 43.8 & 46.0 & 17.6 & 14.7 & 29.3 & 56.0 & +7.0 \\
\rowcolor{yellow!12}
\textbf{TTSR (ours)} & \textbf{60.0} & \textbf{82.4} & \textbf{52.9} & \textbf{45.3} & \textbf{25.6} & \textbf{17.6} & \textbf{34.3} & \textbf{60.8} & \textbf{+11.7} \\
\midrule
\multicolumn{10}{@{}l}{\cellcolor{cyan!6}\textit{Qwen3-8B-Base}} \\
Base Model     & 52.5 & 77.8 & 39.7 & 41.2 & 15.9 & 9.8 & 33.3 & 58.6 & -- \\
SC & 57.5 & 81.8 & 44.1 & 43.8 & 19.9 & 12.3 & 34.8 & 60.6 & +3.3 \\
R-Zero         & 57.5 & 82.0 & 47.8 & 48.7 & 22.6 & 13.4 & 36.9 & 61.5 & +5.2 \\
TTRL           & 62.5 & 84.2 & 50.4 & 50.7 & 26.1 & 15.7 & 38.4 & 62.8 & +7.8 \\
\rowcolor{yellow!12}
\textbf{TTSR (ours)} & \textbf{67.5} & \textbf{87.4} & \textbf{54.8} & \textbf{55.2} & \textbf{30.8} & \textbf{19.1} & \textbf{42.4} & \textbf{66.7} & \textbf{+11.9} \\
\midrule
\multicolumn{10}{@{}l}{\cellcolor{cyan!6}\textit{OctoThinker-8B-Hybrid-Base}} \\
Base Model     & 27.5 & 45.6 & 15.1 & 16.8 & 7.1 & 3.9 & 15.2 & 26.8 & -- \\
SC & 32.5 & 48.6 & 17.3 & 18.2 & 8.6 & 4.9 & 16.2 & 28.3 & +2.1 \\
R-Zero         & 35.0 & 54.0 & 22.8 & 25.4 & 11.6 & 6.8 & 19.7 & 31.4 & +6.1 \\
TTRL           & 37.5 & 57.2 & 26.5 & 28.6 & 13.9 & 8.2 & 21.7 & 33.2 & +8.6 \\
\rowcolor{yellow!12}
\textbf{TTSR (ours)} & \textbf{47.5} & \textbf{64.8} & \textbf{34.2} & \textbf{36.8} & \textbf{19.7} & \textbf{12.4} & \textbf{27.8} & \textbf{39.8} & \textbf{+15.6} \\
\bottomrule
\end{tabular}
\caption{Comparison with Self-Consistency (SC) baseline. SC applies majority voting over $G\!=\!16$ sampled trajectories without parameter updates.}
\label{tab:sc_comparison}
\end{table*}

Self-Consistency provides moderate gains over greedy decoding (Base Model) by leveraging sampling diversity: $+4.2$ on Qwen3-4B, $+3.3$ on Qwen3-8B, and $+2.1$ on OctoThinker-8B.
However, all test-time training methods substantially outperform SC.
Even R-Zero, the weakest TTT baseline, surpasses SC on average, and TTSR's margin over SC is $+7.5$, $+8.6$, and $+13.5$ for the three models respectively.
These results confirm that TTSR's improvements stem from genuine parameter adaptation guided by reflection and variant synthesis, rather than from the additional inference-time computation inherent in sampling multiple trajectories.

\section{Full Component-Level Ablation}
\label{app:ablation_full}

Table~\ref{tab:ablation_full} complements the additive analysis in Table~\ref{tab:ablation} (main text) by removing each component individually from the full TTSR system on Qwen3-8B-Base.

\begin{table*}[t]
\centering
\setlength{\tabcolsep}{5pt}
\renewcommand{\arraystretch}{1.15}
\begin{tabular}{lllll}
\toprule
\textbf{Settings} & \textbf{MATH500} & \textbf{AIME25} & \textbf{Olympiad} & \textbf{GPQA-D} \\
\midrule
\textbf{TTSR (Full)}
& 87.4\:{\textemdash}
& 19.1\:{\textemdash}
& 55.2\:{\textemdash}
& 42.4\:{\textemdash} \\

w/o Reflection-Guided Synthesis
& 85.0 {\color{red}$\downarrow$\,2.4}
& 14.2 {\color{red}$\downarrow$\,4.9}
& 49.9 {\color{red}$\downarrow$\,5.3}
& 38.4 {\color{red}$\downarrow$\,4.0} \\

w/o Teacher Test-Time Update
& 82.8 {\color{red}$\downarrow$\,4.6}
& 13.3 {\color{red}$\downarrow$\,5.8}
& 49.4 {\color{red}$\downarrow$\,5.8}
& 37.9 {\color{red}$\downarrow$\,4.5} \\

w/o Difficulty Reward
& 85.2 {\color{red}$\downarrow$\,2.2}
& 15.4 {\color{red}$\downarrow$\,3.7}
& 51.8 {\color{red}$\downarrow$\,3.4}
& 39.9 {\color{red}$\downarrow$\,2.5} \\

w/o Similarity Penalty
& 86.0 {\color{red}$\downarrow$\,1.4}
& 16.4 {\color{red}$\downarrow$\,2.7}
& 52.8 {\color{red}$\downarrow$\,2.4}
& 40.4 {\color{red}$\downarrow$\,2.0} \\

w/o Strategy Note
& 86.4 {\color{red}$\downarrow$\,1.0}
& 17.2 {\color{red}$\downarrow$\,1.9}
& 53.6 {\color{red}$\downarrow$\,1.6}
& 41.4 {\color{red}$\downarrow$\,1.0} \\

\bottomrule
\end{tabular}
\caption{Full component-level ablation on Qwen3-8B-Base. Each row removes one component from the full TTSR system. Red arrows indicate absolute drops from the full model.}
\label{tab:ablation_full}
\end{table*}

Averaging absolute drops across benchmarks, the individual components rank as: Teacher co-evolution ($-5.2$) $>$ reflection-guided synthesis ($-4.2$) $>$ difficulty reward ($-3.0$) $>$ similarity penalty ($-2.1$) $>$ strategy note ($-1.4$).
The two largest contributors, co-evolution and reflection, jointly define TTSR's core curriculum mechanism, while the remaining components provide complementary but individually smaller gains.

\paragraph{Compound ablation: Reflection + Strategy Note Only.}
Table~\ref{tab:ablation_compound} evaluates a reduced configuration that retains only the reflection mechanism and strategy note.

\begin{table*}[h]
\centering
\setlength{\tabcolsep}{5pt}
\renewcommand{\arraystretch}{1.15}
\begin{tabular}{lllll}
\toprule
\textbf{Configuration} & \textbf{MATH500} & \textbf{AIME25} & \textbf{Olympiad} & \textbf{GPQA-D} \\
\midrule
Base Model
& 77.8
& 9.8
& 41.2
& 33.3 \\

Reflection + Strategy Note Only
& 84.4
& 13.8
& 50.0
& 38.4 \\

\textbf{TTSR (Full)}
& \textbf{87.4}
& \textbf{19.1}
& \textbf{55.2}
& \textbf{42.4} \\

\bottomrule
\end{tabular}
\caption{Compound ablation on Qwen3-8B-Base: retaining only reflection-guided synthesis and strategy note while removing difficulty reward, similarity penalty, and weakness memory.}
\label{tab:ablation_compound}
\end{table*}

Reflection and strategy note alone recover a majority of TTSR's total gain on easier benchmarks (e.g., $69\%$ on MATH500, $63\%$ on Olympiad) but a smaller fraction on harder ones ($43\%$ on AIME25), suggesting that the omitted auxiliary components become increasingly important as task difficulty rises.
The remaining $+3.0$ to $+5.3$ point gap confirms that the omitted auxiliary components substantially amplify the curriculum's effectiveness.

\section{Computational Cost Analysis}
\label{app:compute}

We analyze the per-iteration computational overhead of TTSR relative to TTRL on
Qwen3-4B-Base under the same runtime setup as the main experiments
($M\!=\!2$, $G\!=\!16$, $T\!=\!10$).

\paragraph{Per-iteration breakdown.}
Table~\ref{tab:compute_breakdown} reports cost using a unified normalization:
one TTRL iteration is set to $1.0\times$.
The dominant overhead in TTSR comes from (i) Student rollouts on synthesized
variants and (ii) larger GRPO updates over the mixed batch.
Teacher reflection/synthesis are single-pass generations and contribute
relatively little overhead.
Overall, one TTSR iteration is about $3.0\times$ one TTRL iteration.

\begin{table*}[t]
\centering
\small
\begin{tabular}{@{}lcc@{}}
\toprule
\textbf{Component} & \textbf{TTRL} & \textbf{TTSR (relative)} \\
\midrule
Base TTRL iteration & $1.0\times$ & $1.0\times$ \\
Extra Student rollouts on variants ($M\!=\!2$) & --- & $+1.0\times$ \\
Extra GRPO on mixed batch ($(1+M)\!=\!3$) & --- & $+0.9\times$ \\
Teacher Reflection & --- & $+0.05\times$ \\
Teacher Synthesis & --- & $+0.08\times$ \\
Memory \& Note Update & --- & negligible \\
\midrule
\textbf{Total per iteration} & $\mathbf{1.0\times}$ & $\mathbf{\sim\!3.0\times}$ \\
\bottomrule
\end{tabular}
\caption{Per-iteration computational breakdown on Qwen3-4B-Base. Normalization:
one TTRL iteration $=1.0\times$.}
\label{tab:compute_breakdown}
\end{table*}

\paragraph{Iso-compute comparison.}
Since each TTSR iteration costs $\sim\!3\times$ that of TTRL, running TTSR for
$T\!=\!10$ iterations consumes roughly the same budget as TTRL for $30$
iterations.
Table~\ref{tab:iso_compute} reports this comparison on all eight benchmarks in the main text (Qwen3-4B-Base).

\begin{table*}[t]
\centering
\scriptsize
\setlength{\tabcolsep}{3pt}
\begin{tabular}{@{}lcc|cccccccc|c@{}}
\toprule
\textbf{Method} & \textbf{Iter.} & \textbf{Rel.\ Cost} &
\textbf{AMC23} & \textbf{MATH500} & \textbf{Minerva} & \textbf{Olympiad} & \textbf{AIME24} & \textbf{AIME25} & \textbf{GPQA-D} & \textbf{MMLU-Pro} &
\textbf{Avg} \\
\midrule
Base Model      & --- & $1\times$ &
45.0 & 72.0 & 32.4 & 40.2 & 12.4 & 5.8 & 25.8 & 52.0 &
35.7 \\
TTRL            & 10  & $10\times$ &
55.0 & 79.0 & 43.8 & 46.0 & 17.6 & 14.7 & 29.3 & 56.0 &
42.7 \\
TTRL (extended) & 30  & $30\times$ &
57.5 & 81.0 & 44.1 & 46.4 & 17.8 & 14.8 & 29.8 & 57.4 &
43.6 \\
\textbf{TTSR}   & 10  & $\sim\!30\times$ &
\textbf{60.0} & \textbf{82.4} & \textbf{52.9} & 45.3 & \textbf{25.6} & \textbf{17.6} & \textbf{34.3} & \textbf{60.8} &
\textbf{47.4} \\
\bottomrule
\end{tabular}
\caption{Iso-compute comparison on Qwen3-4B-Base across all eight benchmarks in the main text. TTRL (extended) runs for 30 iterations to approximately match TTSR's total compute.}
\label{tab:iso_compute}
\end{table*}

Extending TTRL from 10 to 30 iterations yields only $+0.9$ additional points
($42.7 \!\to\! 43.6$), indicating clear saturation.
Under the same compute budget, TTSR achieves $47.4$ ($+3.8$ over extended TTRL), confirming that the gains stem from the reflection-guided curriculum rather than additional compute.
The training dynamics in Figure~\ref{fig:training_dynamics} corroborate this: TTRL's accuracy oscillates in later iterations without sustained improvement, whereas TTSR continues to make progress through targeted variant synthesis.

\section{Reflection Quality Evaluation}
\label{app:reflection_eval}

To assess whether the Teacher's reflections accurately identify genuine reasoning weaknesses, we conduct an automated evaluation using GPT-5.2 as an external judge on Qwen3-4B-Base.

\paragraph{Setup.}
We randomly sample 100 reflection instances from MATH500 and 30 from AIME25.
Each instance consists of the original question, the Student's failed reasoning trace, and the Teacher's extracted weakness descriptor (including weakness summary, trigger conditions, and failure signature).
GPT-5.2 evaluates each instance along three dimensions:
\begin{itemize}
    \item \textbf{Weakness Accuracy}: Does the extracted weakness correctly identify the actual reasoning error in the Student's trace?
    \item \textbf{Actionability}: Is the weakness descriptor specific enough to guide targeted question synthesis?
    \item \textbf{Variant Alignment}: Does the subsequently synthesized variant question actually target the identified weakness?
\end{itemize}
Each dimension is scored as \emph{yes} or \emph{no}. We report the fraction of positive judgments.

\begin{table*}[t]
\centering
\small
\begin{tabular}{@{}lccc@{}}
\toprule
\textbf{Benchmark} & \textbf{Accuracy} & \textbf{Actionability} & \textbf{Alignment} \\
\midrule
MATH500 ($n\!=\!100$) & 82.0\% & 88.0\% & 79.0\% \\
AIME25 ($n\!=\!30$)  & 36.7\% & 46.7\% & 33.3\% \\
\bottomrule
\end{tabular}
\caption{GPT-5.2 (2025-12-11) evaluation of Teacher reflection quality on Qwen3-4B-Base.}
\label{tab:reflection_eval}
\end{table*}

On MATH500, the Teacher correctly identifies the reasoning weakness in $82\%$ of cases, and the subsequent variant targets this weakness $79\%$ of the time.
Scores are substantially lower on AIME25: weakness accuracy falls to $36.7\%$, variant alignment falls to $33.3\%$, and actionability reaches $46.7\%$.
This gap is expected given the higher difficulty: competition-level problems involve longer reasoning chains where pinpointing the first error is harder.
The actionability scores ($88.0\%$ and $46.7\%$) remain higher than accuracy on both benchmarks, indicating that even when the weakness diagnosis is imprecise, the descriptor can still provide structural information for variant synthesis.
These results support the ablation finding that reflection-guided synthesis outperforms untargeted variant generation.

\end{document}